\documentclass{article}

\usepackage{microtype}
\usepackage{graphicx}
\usepackage{subcaption}
\usepackage{booktabs} 
\usepackage{float}

\usepackage{hyperref}

\usepackage[accepted]{icml2026}

\usepackage{amsmath}
\usepackage{amssymb}
\usepackage{mathtools}
\usepackage{amsthm}

\usepackage{multirow}
\usepackage{graphicx}
\usepackage{subcaption}
\usepackage{enumitem} 
\usepackage{amsmath}
\usepackage{amssymb}
\setlist[itemize]{leftmargin=5mm}
\usepackage{amsmath, amssymb, mathtools, nccmath}

\usepackage{adjustbox}
\usepackage{float}
\usepackage{tabularray}
\usepackage{tabularx}
\usepackage{sidecap}
\usepackage{comment}
\usepackage{wrapfig}
\usepackage{booktabs,tabularx, colortbl} 
\usepackage{lipsum}
\usepackage[dvipsnames]{xcolor}
\usepackage{amsmath, amsthm, amssymb}
\usepackage{graphicx}
\usepackage{hyperref}
\usepackage{soul}

\theoremstyle{plain}
\newtheorem{theorem}{Theorem}[section]
\newtheorem{proposition}[theorem]{Proposition}
\newtheorem{lemma}[theorem]{Lemma}
\newtheorem{corollary}[theorem]{Corollary}
\theoremstyle{definition}
\newtheorem{definition}[theorem]{Definition}
\newtheorem{assumption}[theorem]{Assumption}
\theoremstyle{remark}
\newtheorem{remark}[theorem]{Remark}

\usepackage{colortbl}
\usepackage{makecell}
\usepackage{tablefootnote}

\definecolor{mintbg}{rgb}{.63,.79,.95}
\colorlet{lightmintbg}{mintbg!50}
\colorlet{lightlightgray}{lightgray!40}
\colorlet{lightyellow}{yellow!40}
\colorlet{lightgreen}{LimeGreen!40}

\let\svthefootnote\thefootnote
\newcommand\freefootnote[1]{%
  \let\thefootnote\relax%
  \footnotetext{#1}%
  \let\thefootnote\svthefootnote%
}

\newcommand{\kv}{KV-Cache}

\newcommand{\xKV}{\texttt{xKV}} 

\newcommand{\xKVSR}{\texttt{xKV-SR}}
\newcommand{\xKSR}{\texttt{xK-SR}}

\usepackage{graphicx}

\usepackage[capitalize,noabbrev]{cleveref}

\icmltitlerunning{xKV: Cross-Layer KV-Cache Compression via Aligned Singular Vector Extraction}

\begin{document}

\twocolumn[
  \icmltitle{xKV: Cross-Layer KV-Cache Compression via Aligned Singular Vector Extraction}



  \icmlsetsymbol{equal}{*}

  \begin{icmlauthorlist}
    \icmlauthor{Chi-Chih Chang}{equal,cornell}
    \icmlauthor{Wei-Cheng Lin}{equal,nycu}
    \icmlauthor{Chien-Yu Lin}{uw}
    \icmlauthor{Hung-Yueh Chiang}{ut}
    \icmlauthor{Yash Akhauri}{cornell}
    \icmlauthor{Yucheng Li}{uos}
    \icmlauthor{Huiqiang Jiang}{msra}
    \icmlauthor{Xilai Dai}{cornell}
    \icmlauthor{Luis Ceze}{uw}
    \icmlauthor{Kai-Chiang Wu}{nycu}
    \icmlauthor{Mohamed Abdelfattah}{cornell}
  \end{icmlauthorlist}

  \icmlaffiliation{cornell}{Cornell University}
  \icmlaffiliation{uw}{University of Washington}
  \icmlaffiliation{nycu}{Department of Computer Science, National Yang Ming Chiao Tung University}
\icmlaffiliation{ut}{The University of Texas at Austin}
\icmlaffiliation{uos}{University of Surrey}
\icmlaffiliation{msra}{Microsoft Research Asia}
  \icmlcorrespondingauthor{Chi-Chih Chang}{cc2869@cornell.edu}

  \icmlkeywords{Machine Learning, ICML}

  \vskip 0.3in
]



\printAffiliationsAndNotice{\icmlEqualContribution}


\begin{abstract}

Long-context Large Language Models (LLMs) enable powerful applications but incur high memory costs due to the key-value states (\kv{}). 
Recent studies attempt to share \kv{} across layers, but these approaches either require expensive pretraining or rely on per-token cross-layer cosine similarity that is often limited in practice. 
We show, via Centered Kernel Alignment (CKA), that the dominant singular vectors of \kv{} are well aligned across layers. 
Motivated by this observation, we propose \textbf{\xKV{}}, a post-training compression method that jointly factorizes grouped-layer \kv{} into a shared low-rank subspace, substantially reducing \kv{} memory.

Across widely used LLMs, \xKV{} achieves up to 8$\times$ \kv{} compression while preserving accuracy on long-context tasks and in multi-turn settings.
To further improve efficiency, we introduce \emph{Selective Reconstruction} (SR) at decode time. 
Combined with SR, \xKV{} achieves up to 4.23$\times$ end-to-end speedup over the full attention baseline, and surpasses notable baselines with 30\% higher throughput under a similar accuracy level.  Overall, \xKV{} provides a plug-and-play approach to reduce both memory and latency for long-context LLM inference. Our code is publicly available at: \url{https://github.com/abdelfattah-lab/xKV}.
\end{abstract} 
\section{Introduction}
Large language models (LLMs) \citep{llama2, openai_gpt-4_2024, team2024gemini, llama3, mistral, anthropic2023claude} have revolutionized numerous artificial intelligence (AI) applications with advanced cognitive capabilities that were previously unattainable with conventional machine learning (ML) models. Recent efforts to extend the context lengths of LLMs have further expanded their potential: open-sourced models now support up to 1M tokens \citep{gradientlongcontextllama3, yang2025qwen251mtechnicalreport}, and proprietary ones like Gemini push this limit even further to 10M tokens \citep{team2024gemini}. 
These extended context windows unlock a wide range of previously impractical applications, such as large-scale information retrieval and debugging or extending a large-scale codebase \citep{deepseek-ai_deepseek-r1_2025, grattafiori2024llama3herdmodels, yang2025qwen251mtechnicalreport, openai_gpt-4_2024}.

\begin{figure*}
\begin{center}
    \includegraphics[width=1.0\textwidth]{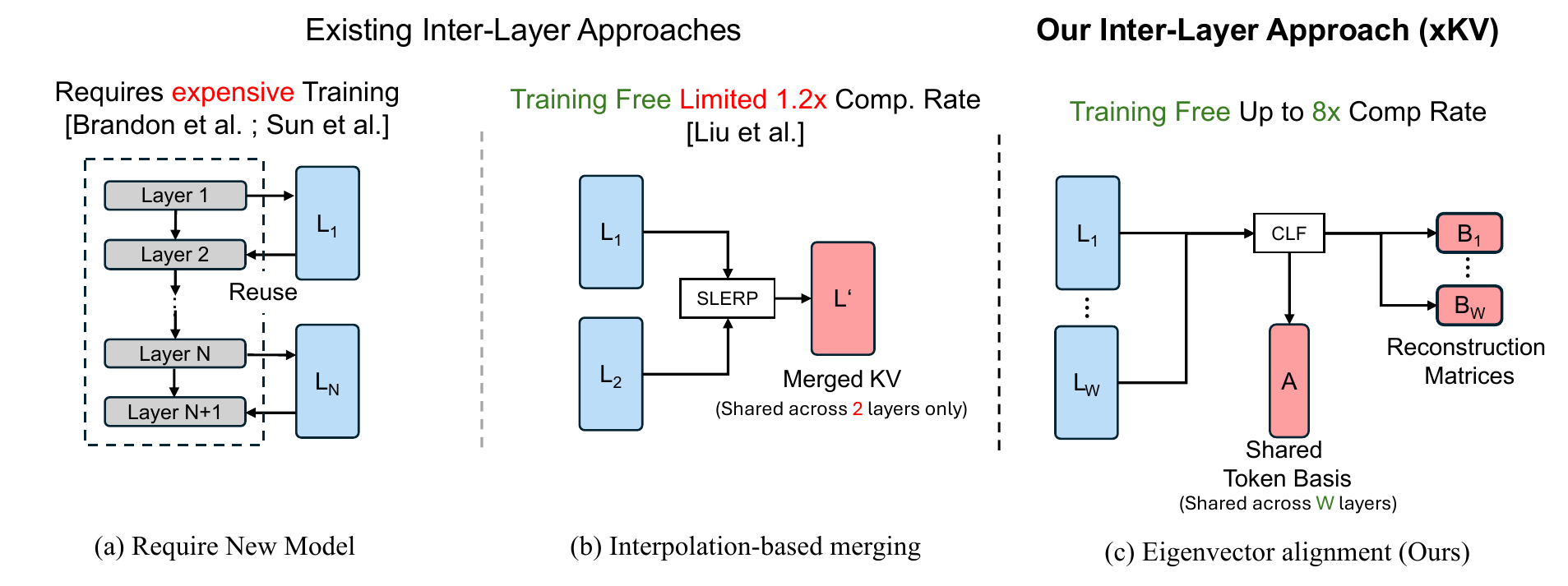}
\end{center}
\caption{\textbf{Comparison of different cross-layer \kv{} compression strategies.} Unlike prior intra-layer methods that overlook inter-layer redundancy or existing inter-layer approaches that require expensive training or offer limited compression rates, \xKV{} introduces a new dimension of inter-layer exploitation. By discovering that dominant singular vectors are highly aligned across layers, \xKV{} extracts a shared token basis through cross-layer factorization (CLF). This training-free, plug-and-play approach enable effective information sharing of $W$ layers ($W>2$), achieving up to 8x compression while maintaining high accuracy on long-context tasks.}
\label{fig: teaser}
\end{figure*}

However, this expanded capability on long-context introduces significant challenges, particularly in the management of key-value (KV) caches during inference \citep{LongContextChallenge, li2024survey}. Typically, KV states are cached to avoid redundant computations; yet, under extended context lengths, the memory consumption of \kv{} rapidly becomes prohibitive. This inflated memory footprint severely limits the number of concurrent inference requests, causing substantial throughput reduction.
To address this, researchers have proposed various approaches to mitigate the large memory footprint of \kv{}s. These include quantization \citep{kvquant, kivi,chen2025powernegativezerodatatype, atom}, token eviction \citep{keyformer, fastgen, streamingllm, h2o, SnapKV, cai2024pyramidkvdynamickvcache,kim2025kvzip}, and low-rank decomposition \citep{sun2024shadowkvkvcacheshadows, chang2025palu, zhang2024lorc, asvd}.
These approaches have primarily focused on intra-layer redundancies that compress the \kv{} of each layer separately. While this can yield respectable per-layer compression, these methods do not utilize potential redundancy across layers.

To exploit cross-layer redundancy \citep{gromov2024unreasonable}, two main lines of work have emerged, as illustrated in Figure~\ref{fig: teaser}. The first, represented by Cross-Layer Attention (CLA) \citep{brandon2024reducing} and YOCO \citep{sun2024you}, introduces new architectures that share a single set of \kv{} across groups of adjacent layers. While effective, these methods require architectural modifications and thus expensive pretraining from scratch, limiting their applicability to existing pretrained models.
A second direction, exemplified by MiniCache \citep{liu2024minicache}, operates in a post-hoc manner by merging adjacent layers’ \kv{} under the assumption of high cosine similarity, implemented via spherical linear interpolation (SLERP) \citep{slerp}. Our analysis, however, shows that such similarity, though present to some extent, is not consistently strong enough across layers to support robust compression, leading to nontrivial accuracy degradation in practice and limited compression rate (see \S\ref{sec:cosine-sim}).
Together, prior methods are limited either by costly pretraining or by fragile similarity assumptions, motivating the need for a new approach.

We posit that the \kv{} exhibits deep structural redundancy across layers that is obscured by low token-wise similarity. We verify this hypothesis using Centered Kernel Alignment (CKA) \citep{kornblith2019similarity}, which evaluates the alignment of centered Gram matrices to capture the similarity of token--token geometries. Our analysis reveals consistently high CKA scores between adjacent layers. This indicates that, while corresponding \kv{} vectors may diverge in cosine similarity, their dominant singular vectors remain highly aligned (see \S\ref{sec:subspace-sim}). Leveraging this insight, we propose exploiting cross-layer redundancy by factorizing multiple adjacent layers into a shared basis, thereby obtaining a highly compact representation.

Building on this insight, we propose \xKV{}, a fully \textit{plug-and-play} compression method that requires no additional fine-tuning or architectural modifications. \xKV{} simultaneously compresses the \kv{} of multiple layers by extracting a \emph{shared} set of singular vectors through cross-layer factorization (CLF), producing a compact token basis reused across adjacent layers as illustrated in Figure \ref{fig: teaser}.
To further reduce overhead at inference, we introduce \textit{Selective Reconstruction (SR)}: instead of reconstructing all tokens, we selectively reconstruct only those relevant to the query. 
The pairing of cross-layer compression with SR (\xKVSR{}) substantially lowers reconstruction cost while preserving model accuracy, making \xKV{} practical for not only reducing memory foot 
print but also deliver end-to-end generation throughput.

We summarize our core contributions as follows:
\begin{itemize}
\item We reveal that KV-Caches share highly aligned dominant singular vectors across layers, unlocking unexploited inter-layer redundancy.
\item We propose \xKV, a training-free method that extracts the shared basis with cross-layer factorization. 
\item  Across RULER and LongBench, \xKV{} achieves 8$\times$ compression on Llama-3.1, Qwen2.5, and DeepSeek-V2 with $\le 3\%$ accuracy loss.
\item By aggressively compressing \kv{}, our method enables larger batch sizes under the same GPU memory budget, reducing attention latency by up to \textbf{3.6$\times$} and improving end-to-end throughput by up to \textbf{4.23$\times$}.

\end{itemize}
\begin{figure*}[!ht]
  \centering
   \begin{subfigure}[b]{0.32\textwidth}
    \centering
      \includegraphics[width=\linewidth]{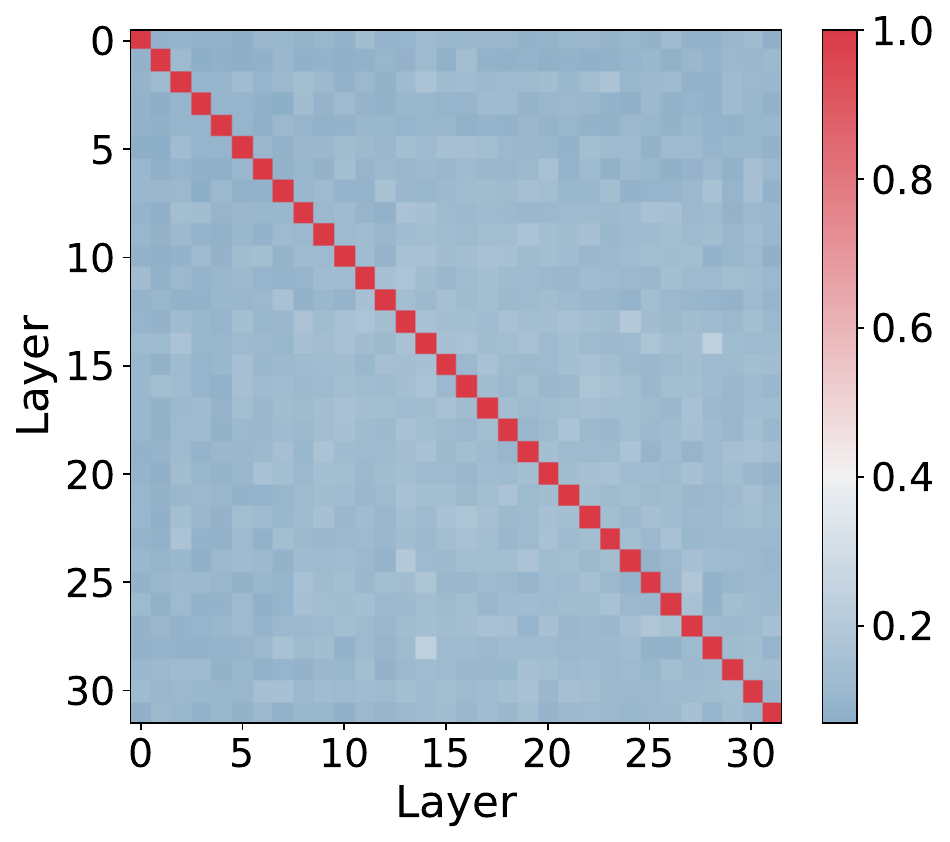}
    \caption{Cosine Similarity Analysis}
    \label{fig:sub1}
  \end{subfigure}
  \hspace{0.02\linewidth}
  \begin{subfigure}[b]{0.32\textwidth}
    \centering
    \includegraphics[width=\linewidth]{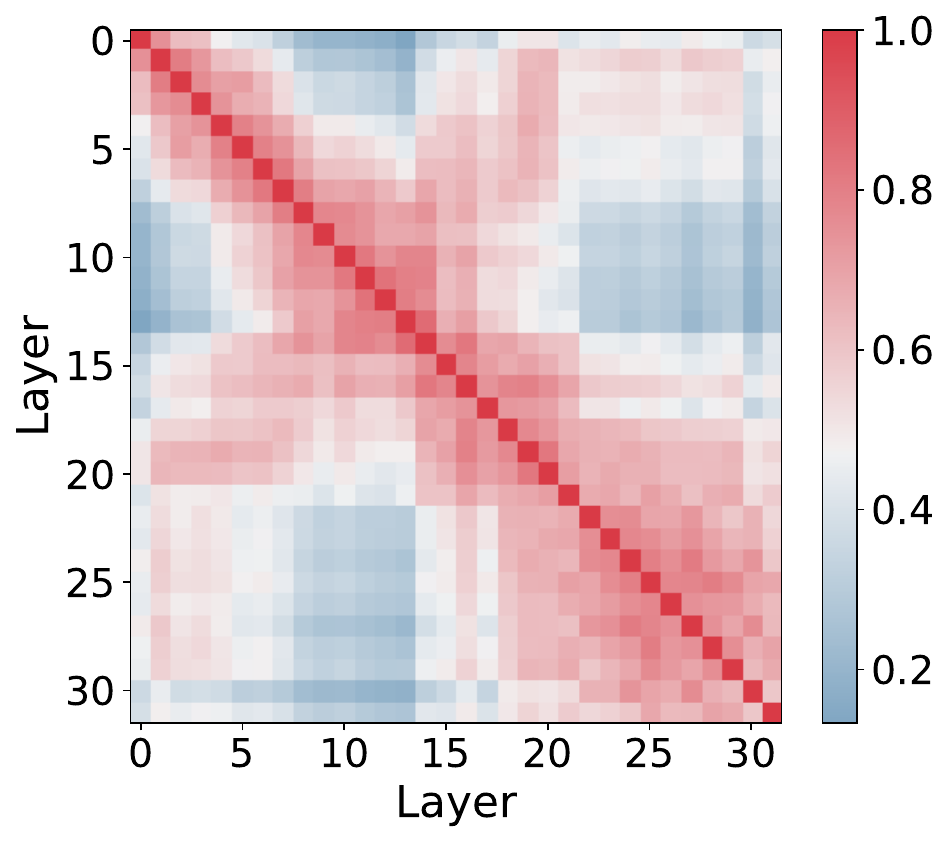}
    \caption{CKA Analysis}
    \label{fig:sub2}
  \end{subfigure}
  \hspace{0.02\linewidth}
  \begin{subfigure}[b]{0.29\textwidth}
    \centering
    \includegraphics[width=\linewidth]{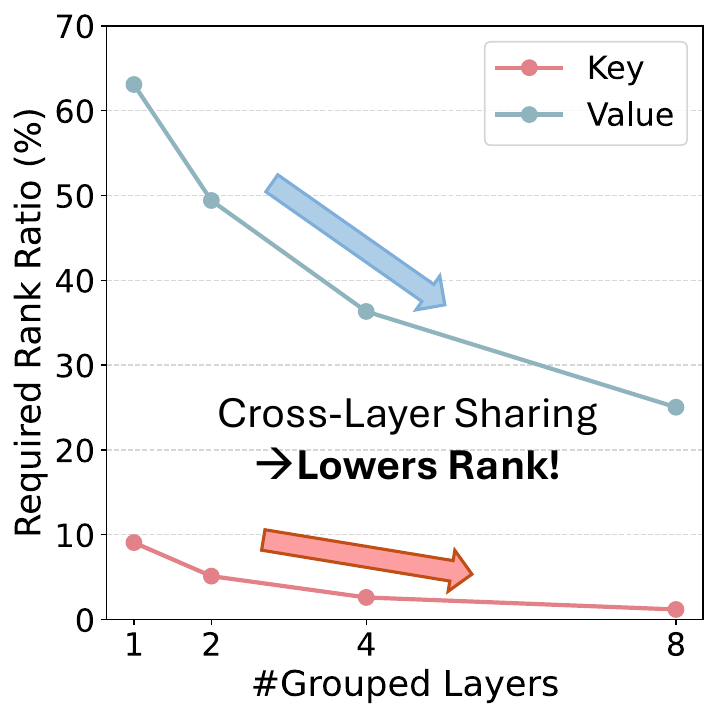}
    \caption{Rank Analysis}
    \label{fig:sub3}
  \end{subfigure}
  \caption{\textbf{(a)} Average Token-wise Cosine Similarity for value-caches across different layers. For each pair of layers, we compute the token-level cosine similarities between their embeddings and average these values into a single similarity score. \textbf{(b)} CKA Matrix for the value-cache. The higher (warmer) values indicate stronger singular vector alignment across layers. \textbf{(c)} Required rank ratio (percentage of total dimension) for capturing 95\% of the cumulative eigenvalues in the key (red) and value (blue) matrices, plotted against the number of grouped layers. For each group, we horizontally concatenate the key/value caches and compute the rank needed to achieve 95\% of the cumulative eigenvalues. As the grouping increases, a smaller rank (relative to total dimension) is required, implying a higher compression rate for the same level of information preservation. We perform these analyses on the \kv{} obtained from Llama-3.1-8B-Instruct, using the multi-valued NIAH dataset from the RULER~\citep{hsieh2024ruler} benchmark.
}
  \label{fig:main}
\end{figure*}

\section{Analysis and Motivation}
\label{sec:motivation}
We begin by examining the cross-layer similarity of \kv{}s with different metrics to reveal the motivation behind the design of \xKV{}. 

\subsection{Notation}
\label{sec:notation}
We consider a Transformer with $N$ decoder blocks and a long prompt of length $L$.
Under GQA, the model has $H_{\mathrm{q}}$ query heads and $H_{\mathrm{kv}}$ KV heads, each with per-head width $d_h$.
We denote the total KV hidden size by $d = H_{\mathrm{kv}}\cdot d_h$, and let $\rho:[H_{\mathrm{q}}]\to[H_{\mathrm{kv}}]$ denote the GQA mapping.

Because our decomposition and reconstruction pipeline applies symmetrically to both keys and values, we present the method for a generic cache
\[
\mathbf{X}_{\ell} \in \mathbb{R}^{L\times d},
\]
which denotes either the \emph{pre-RoPE key cache} or the \emph{value cache}. We use $\mathbf{X}_{\ell,g} \in \mathbb{R}^{L \times d_h}$ to denote the portion of $\mathbf{X}_{\ell}$ corresponding to KV head $g$.

\subsection{Limitations of Token-Wise Similarity} 
\label{sec:cosine-sim}
To understand the constraints of prior art~\citep{liu2024minicache}, we first examine the token-wise cosine similarity across layer pairs. As presented in Figure~\ref{fig:sub1}, adjacent layers exhibit surprisingly low token-level similarity. This observation is critical: it indicates that simplistic merging strategies based on token overlap are fundamentally limited, as they fail to capture the deeper, structural redundancy present in the model.

\subsection{Cross-Layer Alignment with CKA}
\label{sec:subspace-sim}
While token-wise comparisons suggest that adjacent layers are distinct, we posit that their underlying geometric structures remain aligned. To verify this, we adopt Centered Kernel Alignment (CKA)~\citep{kornblith2019similarity} to measure the similarity in the overall structure of two layers’ \kv{}s.

Concretely, for a layer \(\ell\) with \kv{} \(\mathbf{X}_{\ell}\in \mathbb{R}^{L\times d}\), we first define the centered Gram matrix
\[
  \mathbf{G}_{\ell} \;=\; \mathbf{H}\,\mathbf{X}_{\ell}\,\mathbf{X}_{\ell}^\top\,\mathbf{H},
  \quad
  \text{where}
  \quad
  \mathbf{H}\;=\;\mathbf{I}_n \;-\;\tfrac{1}{n}\,\mathbf{1}\,\mathbf{1}^\top.
\]
Then, the \emph{CKA} between two layers \(\ell_1\) and \(\ell_2\) is
\[
  \mathrm{CKA}\bigl(\mathbf{X}_{\ell_1}, \mathbf{X}_{\ell_2}\bigr)
  \;=\;
  \frac{
    \mathrm{trace} \bigl(\mathbf{G}_{\ell_1}\mathbf{G}_{\ell_2}\bigr)
  }{
    \sqrt{\mathrm{trace}\bigl(\mathbf{G}_{\ell_1}^2\bigr)
          \mathrm{trace}\bigl(\mathbf{G}_{\ell_2}^2\bigr)}
  }
  \,.
\]
Unlike token-wise cosine similarity, which compares embeddings point-by-point, CKA captures the similarity between the \emph{entire geometries} of token embeddings. Crucially, a high \(\mathrm{CKA}(\mathbf{X}_{\ell_1}, \mathbf{X}_{\ell_2})\) score implies that the dominant left singular vectors of $\mathbf{X}_{\ell_1}$ are strongly aligned with those of $\ell_2$ (see proof in Appendix~\ref{appendix:cka_proof}). In other words, the basis vectors defining the principal variations in the token space are shared across layers.

\paragraph{Observation: Highly Aligned Basis.} 
Figure~\ref{fig:sub2} visualizes the CKA values between \kv{} layers in Llama-3.1-8B-Instruct. We observe prominent red blocks indicating remarkably high CKA scores between many layer pairs, standing in stark contrast to their modest token-wise cosine similarities. This confirms that although individual token embeddings appear distinct across layers, the dominant singular vectors (\textit{i.e.}, the \emph{basis}) spanning the \kv{} subspaces remain \emph{well-aligned}. Consequently, relying solely on cosine similarity significantly underestimates the potential for \emph{cross-layer} compression. Importantly, this high-CKA phenomenon is not unique to Llama-3.1. We observe the same characteristic singular-vector alignment across diverse model scales, even in a hybrid attention architecture such as GPT-OSS \citep{openai2025gptoss120bgptoss20bmodel}. (See Appendix \ref{appx: more cka}).

\subsection{What Does Highly Align Basis Imply?}

We confirm that the structural alignment observed in Section 2.3  directly translates to compression potential. By examining the spectral properties of KV-Caches concatenated across layers, Figure~\ref{fig:sub3} reveals that the required rank ratio drops significantly as the window size increases. This ratio represents the specific fraction of dimensions needed to preserve 95\% of the cumulative eigenvalues \citep{Jolliffe2002Principal}.

This trend validates our core hypothesis: a single, \textbf{shared token basis} can effectively approximate the collective KV-Caches of multiple adjacent layers. By eliminating the redundancy of storing nearly identical independent bases for each layer, joint cross-layer compression provides a significantly more compact representation than traditional, isolated layer-wise approaches. These findings establish the empirical foundation for xKV, detailed in Section (\S\ref{sec:method}).
\section{Methodology: \xKV{}}
\label{sec:method}

\subsection{Cross-Layer Factorization (CLF)}
\label{sec:cls}

Building upon our observation that the dominant left singular vectors of \kv{}s are well-aligned across adjacent layers (\S~3), we partition the model's $N$ layers into contiguous windows of size $W$. We denote the group of layer indexes in the $k$-th group as $\mathcal{W}_k = \{kW, \ldots, kW+W-1\}$.

For any group $k$, we horizontally concatenate the caches of all layers $\ell \in \mathcal{W}_k$ and compute a rank-$r$ low-rank factorization using SVD:
\begin{equation}
\label{eq:xkv-svd}
\begin{aligned}
\mathbf{X}^{\mathrm{cat}}_k 
  &= \bigl[\mathbf{X}_{kW},\,\ldots,\,\mathbf{X}_{kW+W-1}\bigr] \\
  &\approx \mathbf{U}_{k}\mathbf{\Sigma}_{k}\mathbf{V}_{k}^{\top} \\
  &= \mathbf{A}_k \, \bigl[\mathbf{B}_{kW},\, \ldots,\, \mathbf{B}_{kW+W-1}\bigr],
\end{aligned}
\end{equation}
where $\mathbf{A}_k = \mathbf{U}_{k}\mathbf{\Sigma}_{k} \in \mathbb{R}^{L \times r}$ is the \emph{shared token basis} for the window, and $\mathbf{B}_{\ell} \in \mathbb{R}^{r \times d}$ is the reconstruction matrix specific to layer~$\ell$. Each layer-specific reconstruction matrix $\mathbf{B}_{\ell}$ is composed of $H_{\mathrm{kv}}$ head-specific matrices concatenated column-wise:
\begin{equation}
\mathbf{B}_{\ell} = \bigl[\mathbf{B}_{\ell,1} \;\; \cdots \;\; \mathbf{B}_{\ell,H_{\mathrm{kv}}}\bigr], 
\quad \text{where } \mathbf{B}_{\ell,g} \in \mathbb{R}^{r \times d_h}.
\end{equation}

\subsection{Process During Inference}

\begin{figure}[!thp]
    \centering
    \includegraphics[width=0.9\linewidth]{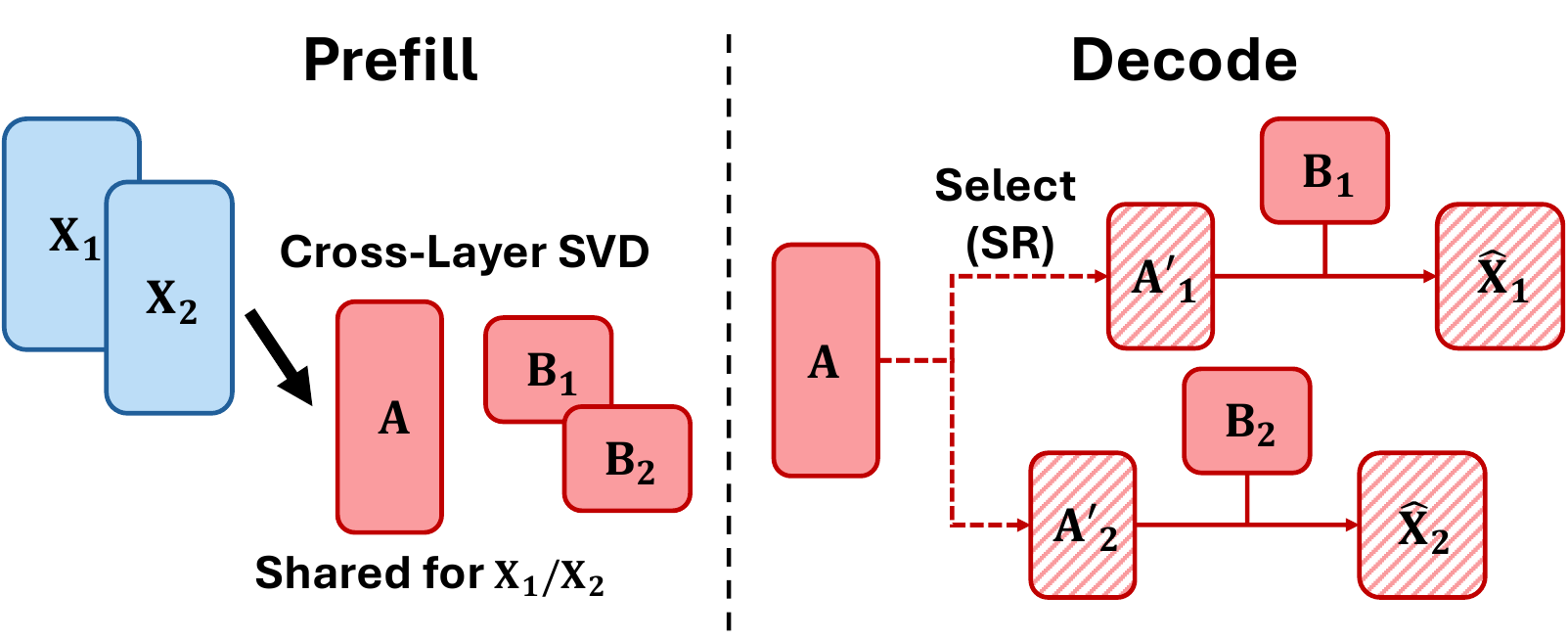}
    \caption{\textbf{Overview of \xKV{} inference.} 
    \textbf{(Left) Prefill:} Cross-layer SVD extracts a shared basis $\mathbf{A}$ for the group and unique reconstruction matrices $\mathbf{B}_{\ell}$ for each layer. We simplify the layer-group index for simplicity. 
    \textbf{(Right) Decode:} Reconstruct from the shared basis. Paired with selective Reconstruction, we reconstruct only the rows corresponding to critical tokens $\mathcal{S}$.}
    \label{fig:overview}
\end{figure}
\vspace{-5pt}

\paragraph{Compression during prefill.}
\label{sec:prefill}
During the prefill phase, we compute Eq.~\eqref{eq:xkv-svd} online to capture prompt dynamics. Empirically, this online decomposition adds negligible overhead (approx.\ 3.9\% of prefill time at 128K context, see Appendix~\ref{App: Prefill SVD Time}).
By sharing $\mathbf{A}_k$ across the window, the storage complexity is reduced from $O(W \cdot L \cdot d)$ to $O(L \cdot r + W \cdot r \cdot d)$. Since $r \ll d$ and $W > 1$, this yields significant memory savings.

\paragraph{Reconstruction during decode.}
For a standard reconstruction of layer $\ell$ (where $k = \lfloor \ell/W \rfloor$), we can compute the full approximation $\widehat{\mathbf{X}}_{\ell}$ for all $L$ tokens:
\[
\widehat{\mathbf{X}}_{\ell} = \mathbf{A}_k\,\mathbf{B}_{\ell}.
\]
This ensures exact recovery of the low-rank approximation but requires a matrix multiplication scaling linearly with sequence length $L$.

\label{sec:sparse}
To address the computational bottleneck of dense reconstruction, we propose \emph{Selective Reconstruction (SR)}. This leverages the native attention sparsity of LLMs \citep{Quest, h2o, sun2024shadowkvkvcacheshadows} by reconstructing only the tokens most relevant to the current query.

At each decoding step $t$, for a specific head $g$ in layer $\ell$, we identify a selected set of token indices $\mathcal{S}_{t,\ell,g} \subseteq [L]$ through approximated attention (See Appendix. \ref{appendix: landmark detail}). We reconstruct only these relevant rows:
\begin{align}
\label{eq:recon-x}
\widehat{\mathbf{X}}_{\ell,g}\bigl[\mathcal{S}_{t,\ell,g}, :\bigr] 
\;=\; 
\mathbf{A}_k\bigl[\mathcal{S}_{t,\ell,g}, :\bigr] \;\mathbf{B}_{\ell,g}.
\end{align}
By fixing the size of this subset such that $|\mathcal{S}| \ll L$, the reconstruction cost becomes constant relative to the sequence length. Note that the subscript $g$ is strictly necessary here because the sparsity pattern $\mathcal{S}$ is head-specific; we must multiply the selected rows of the basis $\mathbf{A}_k$ by the specific columns $\mathbf{B}_{\ell,g}$ corresponding to that head.

\section{Accuracy Evaluations}
\paragraph{Models.} We evaluate \xKV{} on three widely used language models using Grouped-Query Attention (GQA): Llama‐3.1‐8B-Instruct \citep{grattafiori2024llama3herdmodels} (8 KV heads), Qwen2.5-7B-Instruct-1M \citep{yang2025qwen251mtechnicalreport} (4 KV heads), and Qwen3-4B-Instruct-2507 \citep{yang2025qwen3technicalreport} (8 KV heads). 
In Appendix~\ref{appendix: MLA}, we also evaluate \xKV{} on DeepSeek-Coder-V2-Lite-Instruct \citep{dai2024deepseekmoeultimateexpertspecialization} with Multi-head Latent Attention (MLA) and Mixture-of-Experts (MoE) to demonstrate \xKV{}'s high compatibility with emerging efficient Transformer architectures. 

\paragraph{Datasets.}  We select RULER \citep{hsieh2024ruler} as our primary benchmark, which features complex tasks such as retrieval, multi-hop tracking, and question-answering. We also evaluate our approach using Needle In A Haystack (NIAH) \citep{niah} under multi-turn setups. Additionally, we assess performance on LongBench (Appendix~\ref{app: Results on LongBench}).


\paragraph{Baselines.} We compare \xKV{} against a diverse set of state-of-the-art efficiency methods:
\begin{itemize}
    \item \textbf{Token Eviction:} StreamingLLM \citep{streamingllm}, PyramidKV \citep{cai2024pyramidkvdynamickvcache} and SnapKV \citep{SnapKV}.
    \item \textbf{Quantization:} KIVI \citep{kivi}. (2-bit) \footnote{We use a block size of 128 for KIVI-2, yielding an effective bit-width of 2.25 bits/token (${\sim}7.1\times$ compression).}
    \item \textbf{Low-Rank:} Single-SVD (apply SVD on each layer independently)
    \item \textbf{Token Selection \& Hybrid:} Quest \citep{Quest} (dynamic token loading) and ShadowKV \citep{sun2024shadowkvkvcacheshadows} (token selection combined with Single SVD on key and value offloading).
    \item \textbf{Inter-Layer Merging:} MiniCache \citep{liu2024minicache} (based on cross-layer cosine similarity).
\end{itemize}


\paragraph{\xKV{} variants.} 
We evaluate our method under three configurations: \xKV{} applies cross-layer decomposition to both keys and values with dense reconstruction. \xKVSR{} applies selective reconstruction to the compressed keys and values. 
\xKSR{} compresses keys only and offloads values to CPU memory with selective reconstruction; this is for the scenario where GPU memory is still not enough to hold compressed key and value cache (advocated by ShadowKV ~\cite{sun2024shadowkvkvcacheshadows}).

\paragraph{Implementation details.} 
We set the rank for key and value to be 1:1.5 if value compression is applied. 
We use \texttt{torch.svd\_lowrank} API from PyTorch for performing decomposition. We set the cross-layer window size to be 4 as the default setting (see Appendix \ref{subec: Window Size Study}). For baseline, we align MiniCache's official settings to merge half of the layers, from the middle to the end of the LLM, and vary the compression rate by adjusting the layer index at which merging begins. For the token eviction (e.g., SnapKV, PyramidKV) and quantization baseline (KIVI), we adopt the implementation from MInference \citep{jiang2024minference10acceleratingprefilling, li2025scbench} library. For ShadowKV, we use the official open-sourced implementation for evaluation.
We keep the newly generated tokens uncompressed for all comparison targets to ensure fair comparison. Unless specified, we calculate the compression rate by assuming a context length of 64k.

\begin{table*}[!t]
    \centering
    \caption{\kv{} Compression Results: Performance of different methods on the RULER benchmark evaluated at a context length of 64K. \xKV{} consistently achieves a higher accuracy than the Full Attns at the same compression rate or even at a significantly higher compression rate. \textit{Comp.} denote the compression rate. Methods are categorized by Type: \textit{Intra} (Intra-Layer) and \textit{Inter} (Inter-Layer).}
    \resizebox{47em}{!}{
    \begin{tabular}{r|c|c|cccccccccc|c}
    \toprule
    \toprule
    Method & Type & Comp. & N-S1 & N-S2 & N-MK1 & N-MK2 & N-MQ & N-MV & QA-1 & QA-2 & VT & FWE & Avg. \\
    \midrule
    \midrule
    \multicolumn{14}{c}{\textbf{Llama-3.1-8B-Instruct}} \\
    \midrule
    \midrule
    Full Attn      & -- & 1.00 & 100.00 & 100.00 & 98.96 & 97.92 & 98.96 & 97.66 & 83.33 & 59.38 & 97.29 & 85.42 & 91.89 \\
    \midrule
    KIVI-2          & Intra & 7.10 & 100.00 & 96.88  & 98.96 & 90.63 & 91.41 & 89.58 & 80.21 & 55.21 & 81.46 & 84.38 & 86.87 \\
    PyramidKV     & Intra & 8.00 & 100.00 & 100.00 & 100.00 & 96.88 & 100.00 & 98.44 & 83.33 & 57.29 & 95.42 & 68.06 & 89.94 \\
    SnapKV        & Intra & 8.00 & 100.00 & 100.00 & 98.96 & 94.79 & 100.00 & 97.66 & 83.33 & 58.33 & 95.00 & 68.75 & 89.68 \\
    Single SVD & Intra & 8.40 & 25.00 & 51.04 & 61.46 & 96.88 & 28.91 & 44.79 & 47.92 & 36.46 & 3.54 & 61.11 & 45.71 \\
    \midrule
    MiniCache     & Inter & 1.30 & 89.58  & 66.67  & 43.75 & 10.42 & 14.06 & 21.35 & 61.46 & 35.42 & 49.38 & 58.33 & 45.04 \\
    \rowcolor{lightmintbg}
    \xKV{} (Ours) & Inter & 8.03 & 100.00 & 96.88 & 97.92 & 97.92 & 96.09 & 96.62 & 78.13 & 56.25 & 86.67 & 78.47 & 88.50 \\


    \midrule
    \midrule
    \multicolumn{14}{c}{\textbf{Qwen2.5-14B-Instruct-1M}} \\ 
    \midrule
    \midrule
    Full Attn & -- & 1.00 & 100.00 & 100.00 & 100.00 & 98.96 & 100.00 & 98.96 & 80.21 & 64.58 & 99.58 & 91.32 & 93.36 \\
    \midrule
    KIVI-2 & Intra & 6.40 & 100.00 & 98.96 & 95.83 & 88.54 & 98.18 & 92.97 & 78.13 & 64.58 & 97.71 & 89.58 & 90.45 \\
    PyramidKV & Intra & 6.00 & 100.00 & 100.00 & 100.00 & 81.25 & 99.74 & 97.14 & 83.33 & 64.58 & 99.58 & 86.11 & 91.17 \\
    SnapKV & Intra & 6.00 & 100.00 & 100.00 & 100.00 & 84.38 & 99.74 & 95.57 & 81.25 & 65.63 & 99.79 & 90.28 & 91.66 \\
    Single SVD & Intra & 6.35 & 94.79 & 66.67 & 83.33 & 98.96 & 82.81 & 66.15 & 48.96 & 50.00 & 46.04 & 80.21 & 71.79 \\
    \midrule
    MiniCache & Inter & 1.30 & 35.42 & 12.50 & 13.54 & 2.08 & 1.82 & 1.56 & 20.83 & 30.21 & 1.46 & 18.40 & 13.78 \\
    \rowcolor{lightmintbg}xKV-4 (Ours) & Inter & 6.21 & 100.00 & 97.92 & 100.00 & 98.96 & 100.00 & 91.93 & 71.88 & 60.42 & 95.00 & 85.76 & 90.19 \\

    \bottomrule
    \bottomrule
    \end{tabular}
    }
    \label{tab:ruler_comp}
\end{table*}
\subsection{Accuracy and Compression Results.}
\paragraph{Results on RULER Datasets}
Table \ref{tab:ruler_comp} reports the performance of \xKV{} and several representative compression methods on the RULER benchmark at a 64K context length. 
As shown in Table \ref{tab:ruler_comp}, MiniCache suffers dramatic accuracy loss even at a modest 1.3× compression rate. This degradation echos our finding in \S\ref{sec:cosine-sim}), the token-wise cosine similarity in \kv{} across adjacent layers is generally low.
Compared to ShadowKV's compression methods (\textit{i.e.}, single-layer SVD compression), \xKV{} yields substantial accuracy gains: at an 8× compression rate, \xKV{} improves average accuracy by 43\% on Llama-3.1-8B-Instruct and by 8\% on Qwen2.5-7B-Instruct-1M, demonstrating its superior information preservation by exploiting the inherent alignment of \kv{} representations across layers. 

In comparison with token-eviction methods, \xKV{} achieves 88.50\% accuracy on Llama-3.1-8B-Instruct at 8.03× compression, closely matching SnapKV. 
On Qwen2.5-7B-Instruct-1M, however, both SnapKV and Pyramid incur noticeable accuracy degradation. 
We attribute this to Qwen2.5’s inherently more compact KV cache—due to its smaller number of KV heads—which makes information preservation more challenging. 
Despite this, \xKV{} attains 89.22\% average accuracy, narrowing the gap to the non-compressed baseline to just 2.6\%. 
Moreover, \xKV{} surpasses the quantization baseline KIVI-2 by 1.7\% on Llama-3.1-8B while maintaining accuracy on Qwen2.5, where KIVI-2 suffers significant drops. 
Finally, as shown in Appendix~\ref{app: Integrate with Quantization}, our approach can be combined with quantization to further increase compression without sacrificing accuracy.

\begin{figure}[ht]
\centering
\begin{minipage}{0.48\textwidth}
  \centering
  \includegraphics[width=0.8\linewidth]{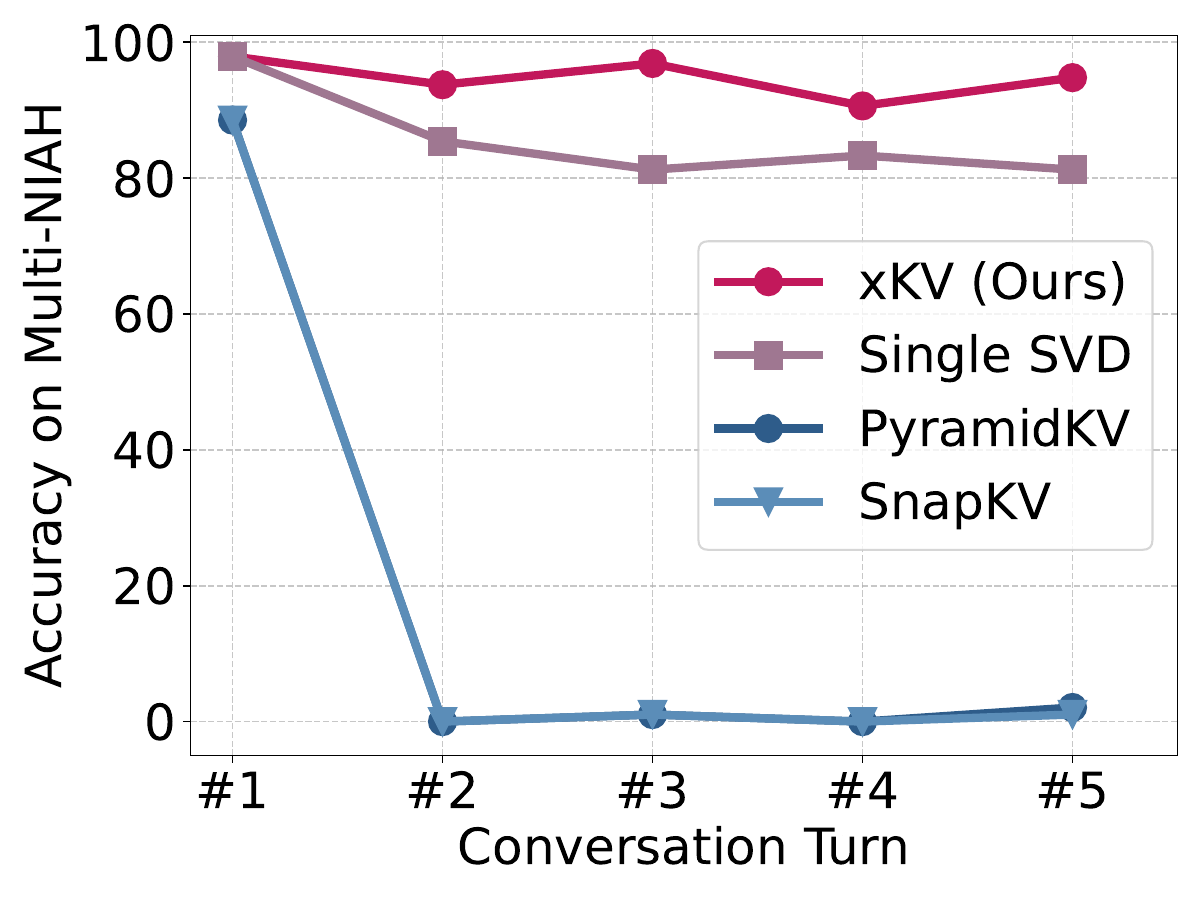} 
  \caption{Accuracy of each conversation turn on Multi-turn NIAH. All compression methods are set at a compression rate of 8$\times$.}
  \label{fig: mt-niah}
\end{minipage}\hfill
\end{figure}

\paragraph{Results on Multi-turn Conversation Datasets.}
\label{para: Results on Multi-turn Conversation Datasets}

We test our method using a multi‐turn Needle‐In‐A‐Haystack (NIAH) benchmark and compare its efficacy against token eviction–based approaches (e.g., SnapKV and PyramidKV). We conduct the evaluation at context length of 64K. Figure \ref{fig: mt-niah} shows results on Llama-3.1-8B-Instruct. SnapKV and PyramidKV both suffer steep declines after the first turn because they evict tokens using the initial attention patterns of the first query and cannot recover context for later queries \citep{li2025scbench}. In contrast, our \xKV{} maintains stable performance across all turns and consistently preserves critical information. 

\begin{table*}[ht]
    \centering
    \caption{\kv{} Compression with Selective Reconstruction Results: Accuracy of different methods on the RULER benchmark at a context length of 64K. Here, "Comp." indicates the total KV-Cache reduction, while the number in parentheses shows the effective GPU memory reduction considering \kv{} offloading. ShadowKV$^\ddagger$ denotes a variant that additionally compresses the value cache.}
    \resizebox{47em}{!}{
    \label{tab:ruler_sparse}
    \begin{tabular}{r|c|cccccccccc|c}
    \toprule
    \toprule
    Method & Comp. & N-S1 & N-S2 & N-MK1 & N-MK2 & N-MQ & N-MV & QA-1 & QA-2 & VT & FWE & Avg. \\
    \midrule
    \multicolumn{13}{c}{\textbf{Llama-3.1-8B-Instruct}} \\
    \midrule
    Full Attn   & 1.00 & 100.00 & 100.00 & 98.96 & 97.92 & 98.96 & 97.66 & 83.33 & 59.38 & 97.29 & 85.42 & 91.89 \\
    Quest      & 1.00 (8.00) & 93.75  & 90.63  & 96.88 & 87.50 & 94.27 & 85.42 & 83.33 & 57.29 & 77.71 & 81.94 & 84.87 \\
    \midrule
    ShadowKV  & 1.64 (9.08) & 100.00 & 100.00 & 98.96 & 97.92 & 96.88 & 94.53 & 82.29 & 60.42 & 66.04 & 74.65 & 87.17 \\
    \rowcolor{lightmintbg}
    \xKSR{} (Ours) & 1.63 (8.90) & 100.00 & 100.00 & 98.96 & 97.92 & 98.44 & 95.31 & 83.33 & 60.42 & 87.92 & 74.65 & 89.70 \\
    \midrule
    ShadowKV$^\ddagger$   & 5.51 & 100.00 & 76.04 & 75.00 & 97.92 & 54.43 & 45.83 & 81.25 & 57.29 & 47.29 & 74.31 & 70.94 \\
    \rowcolor{lightmintbg}
    \xKVSR{} (Ours) & 5.35\footnotemark & 100.00 & 100.00 & 98.96 & 97.92 & 98.44 & 95.57 & 82.29 & 60.42 & 87.29 & 76.04 & 89.69 \\
    \bottomrule
    \bottomrule
    \end{tabular}
    }
\end{table*}

\paragraph{Comparison with Token Selection Methods}
In Table~\ref{tab:ruler_sparse}, we compare \xKSR{}, \xKVSR{}, and two representative token selection baselines, Quest and ShadowKV, using the RULER benchmark at a 64K context length for Llama-3.1-8B-Instruct. For a fair comparison, we fix the token budget (\textit{i.e.}, the number of tokens selected for each decoding step) to be 2k for evaluation targets. Compared with Quest, both \xKSR{} and \xKVSR{} showcase superior accuracy with around 4\% higher in average. As Quest does not entail \kv{} compression but only dynamic loading, it does not reduce the size of the \kv{} and necessitates \kv{} offloading to avoid out-of-memory (OOM). Compared against ShadowKV, \xKSR{} extends its by replacing the single-layer SVD compression key cache with a cross-layer alternative.
At a 1.64× KV-compression rate (8.9× GPU memory reduction considering value offloading), \xKVSR{} closes the accuracy gaps from 4.7\% to around 2.1\%, demonstrating xKV's better capability in preserving information. Leveraging the cross-layer alignment that we observed, \xKVSR{} is able to compress and reduce the \kv{} to a significant 5.35$\times$ while maintaining 89.69\% accuracy, roughly 19\% higher than ShadowKV*. This enables retaining all tensors on GPUs and unlocking the faster inference that avoids the host-device transfer, which improves decoding efficiency over offloading scenarios (See Section~\ref{sec:efficiency}).

\section{Efficiency Studies}
\label{sec:efficiency}

\begin{figure*}[!ht]
    \centering
    \includegraphics[width=0.98\linewidth]{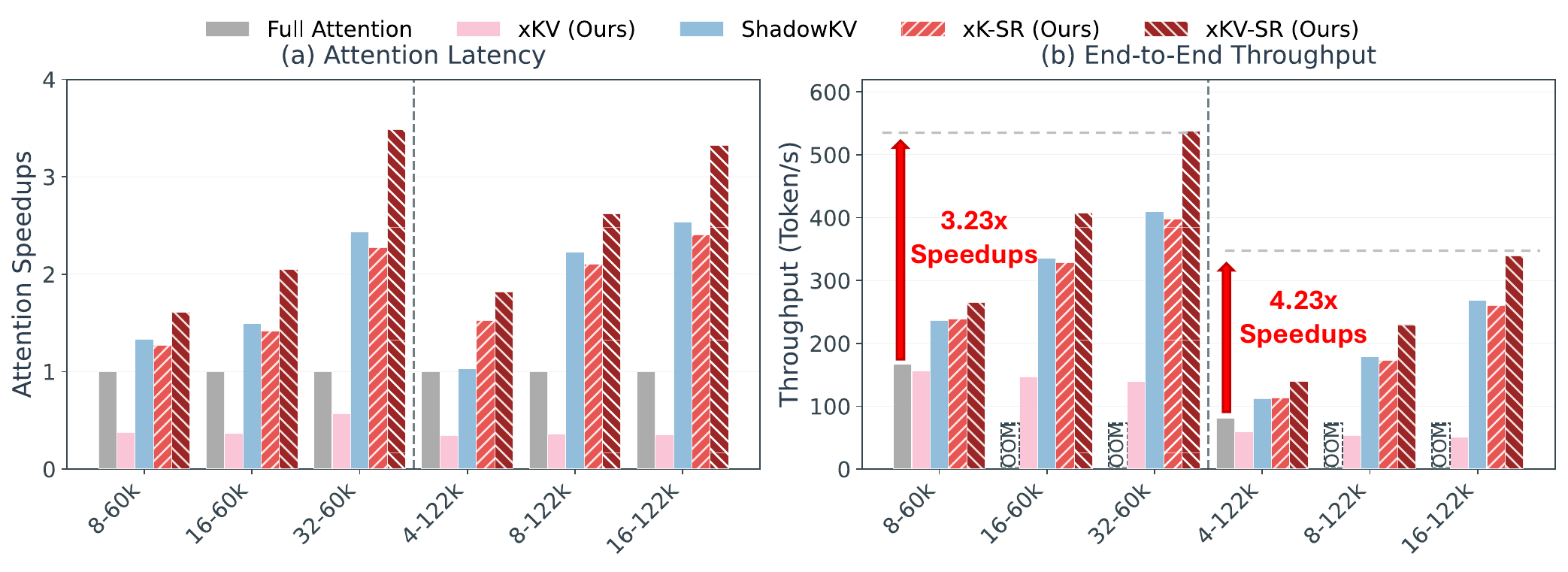}
    \caption{Performance comparison of attention methods on an NVIDIA A100 (80GB) across batch sizes and sequence lengths.
    (a) Attention latency speedup normalized to Full Attention (FlashAttention-2); higher is better.
    (b) End-to-end generation throughput (tokens/s); higher is better. ``OOM'' indicates out-of-memory.
    The dashed vertical line separates 60k and 122k sequence-length settings.}
    \label{fig:efficiency_results}
\end{figure*}

\paragraph{Setup.}
We benchmark Llama-3.1-8B (GQA) on a single NVIDIA A100 GPU (80GB). Figure~\ref{fig:efficiency_results}a reports the attention operation latency speedup normalized to Full Attention (FlashAttention-2), and Figure~\ref{fig:efficiency_results}b reports end-to-end generation throughput (tokens/s) across batch sizes and sequence lengths. Unless stated otherwise, all \xKV{} variants utilize an $8\times$ compressed KV-cache via cross-layer SVD. For ShadowKV and \xKSR{}, we follow the original ShadowKV system, overlapping reconstruction with Value cache transfers from the CPU via PCIe.

\paragraph{Dense Reconstruction.}
Dense reconstruction in \xKV{} successfully removes the KV-cache memory bottleneck of Full Attention, enabling larger-batch inference (e.g., 32-60k) where the baseline suffers from Out-Of-Memory (OOM) errors. However, we observe a computational trade-off: reconstructing all tokens requires additional FLOPs that scale linearly with sequence length $L$, placing reconstruction on the critical path. Consequently, \xKV{} becomes compute-bound, with its normalized kernel speedup staying below $1.0\times$ across all settings and dropping to $\approx 0.4\times$ at $122k$ lengths.

\paragraph{SR Variants (\xKSR{}, \xKVSR{}) \textit{vs}.\ ShadowKV.}
Selective reconstruction (SR) alleviates the compute bottleneck by reconstructing only a sparse, query-relevant subset of tokens, thereby avoiding the $\mathcal{O}(L)$ cost of dense reconstruction. As shown in Figure \ref{fig:efficiency_results}, there are clear gains in latency and throughput compared to \xKV{}.

Next, we look more at the performance difference between \xKSR{} and ShadowKV, which share the same regime with Value cache offloaded and non-compressed. In this case, we observe almost matched throughput across all configurations (Figure 5b) with \xKSR{} slightly falling short. This minor performance gap arises from the increased reconstruction FLOPs inherent in our cross-layer approach. Specifically, at an equivalent memory-saving level, cross-layer factorization uses a larger rank ($r$) for the shared basis across multiple layers than the rank used in isolated single-layer compression. Consequently, reconstructing the cache for each individual layer requires matrix multiplications with larger inner dimensions, slightly increasing the computational overhead per token during the decoding phase (see Appendix. \ref{app:flops-memory}). However, this minor-to-none degradation is justified by accuracy: \xKSR{} yields approximately 2.53\% higher accuracy than ShadowKV by capturing the aligned singular vectors that single-layer methods miss. 

\xKVSR{} achieves the best efficiency by maintaining the compressed cache entirely on the GPU (HBM), bypassing the PCIe ceiling that limits offloading-based methods. It achieves up to a \textbf{3.5$\times$} attention-operation speedup (Figure \ref{fig:efficiency_results}a, 4-122k) and delivers a notable end-to-end throughput improvement of 4.23$\times$ at the 122k context length compared to Full Attention (FlashAttention-2) baseline, providing an approximate \textbf{30\%} throughput improvement over ShadowKV, while simultaneously improving accuracy by 2.5\% (Table \ref{tab:ruler_sparse}).

\footnotetext{The final compression rate accounts for the memory overhead of the landmarks used to compute selective indices. See Appendix~\ref{app:flops-memory} for details.}
\section{Related Works}
\label{sec:related work}

\paragraph{Intra-Layer \kv{} Compression.} 
Most prior approaches focus on reducing the \kv{} size within each layer independently (intra-layer redundancy). Quantization methods \citep{kivi, kvquant} reduce the memory footprint by storing tensors in lower precision (e.g., 2-bit or 4-bit), though often requiring custom kernels to maintain accuracy. 
Token Eviction strategies \citep{streamingllm,SnapKV,cai2024pyramidkvdynamickvcache,h2o} prune less important tokens based on attention scores; while effective, they permanently discard information, which can degrade performance in long-context retrieval tasks. 
Another major direction exploits the low-rank nature of the \kv{}. 
Architectures like Multi-Head Latent Attention (MLA) \citep{deepseekai2024deepseekv2} cache compressed latent representations but require training from scratch. Post-training methods decompose the weight matrices \citep{asvd, chang2025palu, zhang2024lorc} or the \kv{} itself \citep{saxena-etal-2024-eigen, sun2024shadowkvkvcacheshadows} into low-rank forms. 
While these methods yield respectable compression, they treat each layer in isolation, failing to utilize the extra redundancy present across layers.
\vspace{-5pt}
\paragraph{Cross-Layer \kv{} Compression.} 
Going beyond the intra-layer perspective, another stream of research explores inter-layer redundancy of \kv{} \citep{brandon2024reducing, sun2024you, wu2024layer, liu2024minicache, dong2025hymba}. CLA \citep{brandon2024reducing} and YOCO\citep{sun2024you} both modify the Transformer model architecture so that later layers can directly reuse or reference KV states from earlier layers. 
LCKV \citep{wu2024layer} restricts full KV storage to a small subset of layers, foregoing caches in other layers. However, these methods rely on retraining or model fine‐tuning, which makes them less flexible. Minicache \citep{liu2024minicache}, in contrast, provides a flexible post-training alternative by merging the key and value tokens from adjacent similar layers using spherical linear interpolation. Our approach goes further by extracting shared singular vectors of multiple layers' \kv{}s, thereby enabling higher compression.


\paragraph{Dynamic Token Selection and KV Offloading.} 
A complementary line of work accelerates decoding by selecting a small subset of context tokens per step (dynamic sparse attention). 
Quest \citep{Quest} introduced query-aware page selection to reduce attention computation, yet it does not compress the KV-Cache tensors, often necessitating CPU offloading to manage memory footprints. 
ShadowKV \citep{sun2024shadowkvkvcacheshadows} pioneered the concept of \textit{selective reconstruction}, pairing sparse token selection with low-rank compression to reconstruct only critical tokens on-the-fly. 
However, limited by the accuracy constraints of intra-layer SVD, ShadowKV is forced to offload Value states to the CPU, leaving inference speed bounded by PCIe bandwidth. 
Our work closes this loop. We integrate \xKV{}'s cross-layer shared basis into this selective reconstruction pipeline. Unlike ShadowKV, our approach achieves sufficient fidelity to keep both compressed Keys and Values entirely on the GPU. 
Thus, while \xKV{} leverages the selection paradigm pioneered by ShadowKV, it upgrades the underlying compression substrate to eliminate host-device transfers, transforming the system into a faster, fully on-device inference engine.
\section{Limitations and Future Work}

\paragraph{Long Generation Scenario.}
Our study focuses on the long-prefill setting, where only the initial context is compressed while tokens generated during decoding remain uncompressed. This regime covers many long-context applications (e.g., information retrieval \citep{perplexityAI} and database QA), but it does not address test-time scaling under extended generation, which the cumulative \kv{} can also become the bottleneck. We leave to future work how to leverage the observed cross-layer alignment of the KV-cache’s dominant singular vectors and proposed cross-layer SVD to tackle long-generation scenarios.

\section{Conclusion}
We introduce \xKV{}, a plug-and-play compression method for key-value (KV) caches that exploits inter-layer redundancy. Our approach reveals that \kv{}s across different layers share highly aligned basis vectors. Leveraging this property, we apply a cross-layer SVD to compress multiple \kv{}s into a shared low-rank subspace. Experiments demonstrate that \xKV{} outperforms accuracy on all other compression methods, including representative inter-layer approaches and intra-layer methods such as quantization, token eviction, and single-layer SVD. At roughly 8× compression, \xKV{} keeps average accuracy within ~2–3 percentage points of the non-compressed baseline, and it remains robust in multi-turn settings. With \emph{Selective Reconstruction} (SR), our fastest alternative \xKVSR{} reaches up to \textbf{4.23$\times$} faster generation throughputs on A100 GPU, highlighting \xKV{} as a practical approach to reduce both memory footprint and latency for long-context LLM inference.

\vspace{-10pt}
\section*{Impact Statement}
This paper presents work whose goal is to advance the field
of Machine Learning. There are many potential societal
consequences of our work, none which we feel must be
specifically highlighted here.
\vspace{-10pt}
\section*{Acknowledgements} This work was supported in part by the NSF CAREER Grant No. 2339084 and by an Nvidia Research Gift. We thank Zhichen Zeng for the discussions on system design and writing.

\bibliography{example_paper}

@article{llama2,
  title={Llama 2: Open foundation and fine-tuned chat models},
  author={Touvron, Hugo and Martin, Louis and Stone, Kevin and Albert, Peter and Almahairi, Amjad and Babaei, Yasmine and Bashlykov, Nikolay and Batra, Soumya and Bhargava, Prajjwal and Bhosale, Shruti and others},
  journal={arXiv preprint arXiv:2307.09288},
  year={2023}
}

@misc{llama3,
      title={Introducing Meta Llama 3: The most capable openly available LLM to date}, 
      url={https://ai.meta.com/blog/meta-llama-3/},
      author={Meta AI},
      year={2024},
}

@misc{mistral,
      title={Mistral 7B}, 
      author={Albert Q. Jiang and Alexandre Sablayrolles and Arthur Mensch and Chris Bamford and Devendra Singh Chaplot and Diego de las Casas and Florian Bressand and Gianna Lengyel and Guillaume Lample and Lucile Saulnier and Lélio Renard Lavaud and Marie-Anne Lachaux and Pierre Stock and Teven Le Scao and Thibaut Lavril and Thomas Wang and Timothée Lacroix and William El Sayed},
      year={2023},
      eprint={2310.06825},
      archivePrefix={arXiv},
      primaryClass={cs.CL}
}

@article{atom,
  title={Atom: Low-bit quantization for efficient and accurate llm serving},
  author={Zhao, Yilong and Lin, Chien-Yu and Zhu, Kan and Ye, Zihao and Chen, Lequn and Zhenga, Size and Ceze, Luis and Krishnamurthy, Arvind and Chen, Tianqi and Kasikci, Baris},
  journal={arXiv preprint arXiv:2310.19102},
  year={2023}
}

@article{kvquant,
  title={KVQuant: Towards 10 Million Context Length LLM Inference with KV Cache Quantization},
  author={Hooper, Coleman and Kim, Sehoon and Mohammadzadeh, Hiva and Mahoney, Michael W and Shao, Yakun Sophia and Keutzer, Kurt and Gholami, Amir},
  journal={arXiv preprint arXiv:2401.18079},
  year={2024}
}

@article{kivi,
  title={KIVI: A Tuning-Free Asymmetric 2bit Quantization for KV Cache},
  author={Liu, Zirui and Yuan, Jiayi and Jin, Hongye and Zhong, Shaochen and Xu, Zhaozhuo and Braverman, Vladimir and Chen, Beidi and Hu, Xia},
  journal={arXiv preprint arXiv:2402.02750},
  year={2024}
}

@inproceedings{streamingllm,
    title={Efficient Streaming Language Models with Attention Sinks},
    author={Guangxuan Xiao and Yuandong Tian and Beidi Chen and Song Han and Mike Lewis},
    booktitle={The Twelfth International Conference on Learning Representations},
    year={2024},
    url={https://openreview.net/forum?id=NG7sS51zVF}
}

@article{h2o,
  title={H2o: Heavy-hitter oracle for efficient generative inference of large language models},
  author={Zhang, Zhenyu and Sheng, Ying and Zhou, Tianyi and Chen, Tianlong and Zheng, Lianmin and Cai, Ruisi and Song, Zhao and Tian, Yuandong and R{\'e}, Christopher and Barrett, Clark and others},
  journal={Advances in Neural Information Processing Systems},
  volume={36},
  year={2024}
}

@inproceedings{fastgen,
    title={Model Tells You What to Discard: Adaptive {KV} Cache Compression for {LLM}s},
    author={Suyu Ge and Yunan Zhang and Liyuan Liu and Minjia Zhang and Jiawei Han and Jianfeng Gao},
    booktitle={The Twelfth International Conference on Learning Representations},
    year={2024},
    url={https://openreview.net/forum?id=uNrFpDPMyo}
}

@article{keyformer,
  title={Keyformer: KV Cache reduction through key tokens selection for Efficient Generative Inference},
  author={Adnan, Muhammad and Arunkumar, Akhil and Jain, Gaurav and Nair, Prashant and Soloveychik, Ilya and Kamath, Purushotham},
  journal={Proceedings of Machine Learning and Systems},
  volume={7},
  year={2024}
}

@misc{asvd,
      title={ASVD: Activation-aware Singular Value Decomposition for Compressing Large Language Models}, 
      author={Zhihang Yuan and Yuzhang Shang and Yue Song and Qiang Wu and Yan Yan and Guangyu Sun},
      year={2023},
      eprint={2312.05821},
      archivePrefix={arXiv},
      primaryClass={cs.CL}
}

@article{RoPE,
  author       = {Jianlin Su and
                  Yu Lu and
                  Shengfeng Pan and
                  Bo Wen and
                  Yunfeng Liu},
  title        = {RoFormer: Enhanced Transformer with Rotary Position Embedding},
  journal      = {CoRR},
  volume       = {abs/2104.09864},
  year         = {2021},
  url          = {https://arxiv.org/abs/2104.09864},
  eprinttype    = {arXiv},
  eprint       = {2104.09864},
  bibsource    = {dblp computer science bibliography, https://dblp.org}
}

@article{deepseekai2024deepseekv2,
  title={Deepseek-v2: A strong, economical, and efficient mixture-of-experts language model},
  author={Liu, Aixin and Feng, Bei and Wang, Bin and Wang, Bingxuan and Liu, Bo and Zhao, Chenggang and Dengr, Chengqi and Ruan, Chong and Dai, Damai and Guo, Daya and others},
  journal={arXiv preprint arXiv:2405.04434},
  year={2024}
}

@inproceedings{LCC,
  title={Longcoder: A long-range pre-trained language model for code completion},
  author={Guo, Daya and Xu, Canwen and Duan, Nan and Yin, Jian and McAuley, Julian},
  booktitle={International Conference on Machine Learning},
  pages={12098--12107},
  year={2023},
  organization={PMLR}
}

@misc{RepoBench,
      title={RepoBench: Benchmarking Repository-Level Code Auto-Completion Systems}, 
      author={Tianyang Liu and Canwen Xu and Julian McAuley},
      year={2023},
      eprint={2306.03091},
      archivePrefix={arXiv},
      primaryClass={id='cs.CL' full_name='Computation and Language' is_active=True alt_name='cmp-lg' in_archive='cs' is_general=False description='Covers natural language processing. Roughly includes material in ACM Subject Class I.2.7. Note that work on artificial languages (programming languages, logics, formal systems) that does not explicitly address natural-language issues broadly construed (natural-language processing, computational linguistics, speech, text retrieval, etc.) is not appropriate for this area.'}
}

@misc{LongContextChallenge,
      title={Challenges in Deploying Long-Context Transformers: A Theoretical Peak Performance Analysis}, 
      author={Yao Fu},
      year={2024},
      eprint={2405.08944},
      archivePrefix={arXiv},
      primaryClass={cs.LG},
      url={https://arxiv.org/abs/2405.08944}, 
}

@inproceedings{Quest,
 title={{QUEST: Query-Aware Sparsity for Efficient Long-Context LLM Inference}},
 author={Jiaming Tang and Yilong Zhao and Kan Zhu and Guangxuan Xiao and Baris Kasikci and Song Han},
 booktitle={Proceedings of the International Conference on Machine Learning (ICML)},
 year={2024},
}

@inproceedings{SnapKV,
title={Snap{KV}: {LLM} Knows What You are Looking for Before Generation},
author={Yuhong Li and Yingbing Huang and Bowen Yang and Bharat Venkitesh and Acyr Locatelli and Hanchen Ye and Tianle Cai and Patrick Lewis and Deming Chen},
booktitle={The Thirty-eighth Annual Conference on Neural Information Processing Systems},
year={2024},
url={https://openreview.net/forum?id=poE54GOq2l}
}

@article{grattafiori2024llama3herdmodels,
  title={The llama 3 herd of models},
  author={Dubey, Abhimanyu and Jauhri, Abhinav and Pandey, Abhinav and Kadian, Abhishek and Al-Dahle, Ahmad and Letman, Aiesha and Mathur, Akhil and Schelten, Alan and Yang, Amy and Fan, Angela and others},
  journal={arXiv e-prints},
  pages={arXiv--2407},
  year={2024}
}

@article{team2024gemini,
  title={Gemini 1.5: Unlocking multimodal understanding across millions of tokens of context},
  author={Team, Gemini and Georgiev, Petko and Lei, Ving Ian and Burnell, Ryan and Bai, Libin and Gulati, Anmol and Tanzer, Garrett and Vincent, Damien and Pan, Zhufeng and Wang, Shibo and others},
  journal={arXiv preprint arXiv:2403.05530},
  year={2024}
}

@misc{yang2025qwen251mtechnicalreport,
      title={Qwen2.5-1M Technical Report}, 
      author={An Yang and Bowen Yu and Chengyuan Li and Dayiheng Liu and Fei Huang and Haoyan Huang and Jiandong Jiang and Jianhong Tu and Jianwei Zhang and Jingren Zhou and Junyang Lin and Kai Dang and Kexin Yang and Le Yu and Mei Li and Minmin Sun and Qin Zhu and Rui Men and Tao He and Weijia Xu and Wenbiao Yin and Wenyuan Yu and Xiafei Qiu and Xingzhang Ren and Xinlong Yang and Yong Li and Zhiying Xu and Zipeng Zhang},
      year={2025},
      eprint={2501.15383},
      archivePrefix={arXiv},
      primaryClass={cs.CL},
      url={https://arxiv.org/abs/2501.15383}, 
}

@misc{openai_gpt-4_2024,
	title = {{GPT}-4 {Technical} {Report}},
	url = {http://arxiv.org/abs/2303.08774},
	doi = {10.48550/arXiv.2303.08774},
	urldate = {2025-02-03},
	publisher = {arXiv},
	author = {OpenAI and Achiam, Josh and Adler, Steven and Agarwal, Sandhini and Ahmad, Lama and Akkaya, Ilge and Aleman, Florencia Leoni and Almeida, Diogo and Altenschmidt, Janko and Altman, Sam and others},
	month = mar,
	year = {2024},
	note = {arXiv:2303.08774 [cs]},
}

@misc{deepseek-ai_deepseek-r1_2025,
	title = {{DeepSeek}-{R1}: {Incentivizing} {Reasoning} {Capability} in {LLMs} via {Reinforcement} {Learning}},
	copyright = {arXiv.org perpetual, non-exclusive license},
	shorttitle = {{DeepSeek}-{R1}},
	url = {https://arxiv.org/abs/2501.12948},
	doi = {10.48550/ARXIV.2501.12948},
	urldate = {2025-02-03},
	publisher = {arXiv},
	author = {{DeepSeek-AI} and Guo, Daya and Yang, Dejian and Zhang, Haowei and Song, Junxiao and Zhang, Ruoyu and Xu, Runxin and Zhu, Qihao and Ma, Shirong and Wang, Peiyi and others},
	year = {2025},
	note = {Version Number: 1},
}

@inproceedings{
liu2024minicache,
title={MiniCache: {KV} Cache Compression in Depth Dimension for Large Language Models},
author={Akide Liu and Jing Liu and Zizheng Pan and Yefei He and Gholamreza Haffari and Bohan Zhuang},
booktitle={The Thirty-eighth Annual Conference on Neural Information Processing Systems},
year={2024},
url={https://openreview.net/forum?id=sgVOjDqUMT}
}

@inproceedings{kornblith2019similarity,
  title={Similarity of neural network representations revisited},
  author={Kornblith, Simon and Norouzi, Mohammad and Lee, Honglak and Hinton, Geoffrey},
  booktitle={International conference on machine learning},
  pages={3519--3529},
  year={2019},
  organization={PMLR}
}

@inproceedings{
brandon2024reducing,
title={Reducing Transformer Key-Value Cache Size with Cross-Layer Attention},
author={William Brandon and Mayank Mishra and Aniruddha Nrusimha and Rameswar Panda and Jonathan Ragan-Kelley},
booktitle={The Thirty-eighth Annual Conference on Neural Information Processing Systems},
year={2024},
url={https://openreview.net/forum?id=M2UzLRoqic}
}

@inproceedings{li2025scbench,
    title={{SCB}ench: A {KV} Cache-Centric Analysis of Long-Context Methods},
    author={Yucheng Li and Huiqiang Jiang and Qianhui Wu and Xufang Luo and Surin Ahn and Chengruidong Zhang and Amir H. Abdi and Dongsheng Li and Jianfeng Gao and Yuqing Yang and Lili Qiu},
    booktitle={The Thirteenth International Conference on Learning Representations},
    year={2025},
    url={https://openreview.net/forum?id=gkUyYcY1W9}
}

@misc{sun2024shadowkvkvcacheshadows,
      title={ShadowKV: KV Cache in Shadows for High-Throughput Long-Context LLM Inference}, 
      author={Hanshi Sun and Li-Wen Chang and Wenlei Bao and Size Zheng and Ningxin Zheng and Xin Liu and Harry Dong and Yuejie Chi and Beidi Chen},
      year={2024},
      eprint={2410.21465},
      archivePrefix={arXiv},
      primaryClass={cs.LG},
      url={https://arxiv.org/abs/2410.21465}, 
}

@inproceedings{
chang2025palu,
title={Palu: {KV}-Cache Compression with Low-Rank Projection},
author={Chi-Chih Chang and Wei-Cheng Lin and Chien-Yu Lin and Chong-Yan Chen and Yu-Fang Hu and Pei-Shuo Wang and Ning-Chi Huang and Luis Ceze and Mohamed S. Abdelfattah and Kai-Chiang Wu},
booktitle={The Thirteenth International Conference on Learning Representations},
year={2025},
url={https://openreview.net/forum?id=LWMS4pk2vK}
}

@inproceedings{
sun2024you,
title={You Only Cache Once: Decoder-Decoder Architectures for Language Models},
author={Yutao Sun and Li Dong and Yi Zhu and Shaohan Huang and Wenhui Wang and Shuming Ma and Quanlu Zhang and Jianyong Wang and Furu Wei},
booktitle={The Thirty-eighth Annual Conference on Neural Information Processing Systems},
year={2024},
url={https://openreview.net/forum?id=25Ioxw576r}
}

@article{hsieh2024ruler,
  title={RULER: What's the Real Context Size of Your Long-Context Language Models?},
  author={Cheng-Ping Hsieh and Simeng Sun and Samuel Kriman and Shantanu Acharya and Dima Rekesh and Fei Jia and Yang Zhang and Boris Ginsburg},
  year={2024},
  journal={arXiv preprint arXiv:2404.06654},
}

@article{slerp,
author = {Shoemake, Ken},
title = {Animating rotation with quaternion curves},
year = {1985},
issue_date = {Jul. 1985},
publisher = {Association for Computing Machinery},
address = {New York, NY, USA},
volume = {19},
number = {3},
issn = {0097-8930},
url = {https://doi.org/10.1145/325165.325242},
doi = {10.1145/325165.325242},
journal = {SIGGRAPH Comput. Graph.},
month = jul,
pages = {245–254},
numpages = {10}
}

@article{gromov2024unreasonable,
  title={The unreasonable ineffectiveness of the deeper layers},
  author={Gromov, Andrey and Tirumala, Kushal and Shapourian, Hassan and Glorioso, Paolo and Roberts, Daniel A},
  journal={arXiv preprint arXiv:2403.17887},
  year={2024}
}

@article{zhang2024lorc,
  title={Lorc: Low-rank compression for llms kv cache with a progressive compression strategy},
  author={Zhang, Rongzhi and Wang, Kuang and Liu, Liyuan and Wang, Shuohang and Cheng, Hao and Zhang, Chao and Shen, Yelong},
  journal={arXiv preprint arXiv:2410.03111},
  year={2024}
}

@inproceedings{saxena-etal-2024-eigen,
    title = "Eigen Attention: Attention in Low-Rank Space for {KV} Cache Compression",
    author = "Saxena, Utkarsh  and
      Saha, Gobinda  and
      Choudhary, Sakshi  and
      Roy, Kaushik",
    editor = "Al-Onaizan, Yaser  and
      Bansal, Mohit  and
      Chen, Yun-Nung",
    booktitle = "Findings of the Association for Computational Linguistics: EMNLP 2024",
    month = nov,
    year = "2024",
    address = "Miami, Florida, USA",
    publisher = "Association for Computational Linguistics",
    url = "https://aclanthology.org/2024.findings-emnlp.899/",
    doi = "10.18653/v1/2024.findings-emnlp.899",
    pages = "15332--15344"
}

@inproceedings{
dong2025hymba,
title={Hymba: A Hybrid-head Architecture for Small Language Models},
author={Xin Dong and Yonggan Fu and Shizhe Diao and Wonmin Byeon and ZIJIA CHEN and Ameya Sunil Mahabaleshwarkar and Shih-Yang Liu and Matthijs Van keirsbilck and Min-Hung Chen and Yoshi Suhara and Yingyan Celine Lin and Jan Kautz and Pavlo Molchanov},
booktitle={The Thirteenth International Conference on Learning Representations},
year={2025},
url={https://openreview.net/forum?id=A1ztozypga}
}

@inproceedings{wu2024layer,
  title={Layer-condensed kv cache for efficient inference of large language models},
  author={Wu, Haoyi and Tu, Kewei},
  booktitle={Proceedings of the 62nd Annual Meeting of the Association for Computational Linguistics (Volume 1: Long Papers)},
  pages={11175--11188},
  year={2024}
}

@misc{anthropic2023claude,
  author = {Anthropic},
  title = {{Claude}: A Conversational AI Assistant},
  year = {2023},
  note = {Large Language Model. Version 1.0. Accessed: 2025-03-13},
  url = {https://www.anthropic.com/claude}
}

@article{gradientlongcontextllama3,
  title={Llama 3 gradient: A series of long context models, 2024},
  author={Pekelis, Leonid and Feil, Michael and Moret, Forrest and Huang, Mark and Peng, Tiffany},
  journal={URL https://gradient.ai/blog/scaling-rotational-embeddings-for-long-context-language-models},
  year = {2024}
}

@article{li2024survey,
  title={A survey on large language model acceleration based on kv cache management},
  author={Li, Haoyang and Li, Yiming and Tian, Anxin and Tang, Tianhao and Xu, Zhanchao and Chen, Xuejia and Hu, Nicole and Dong, Wei and Li, Qing and Chen, Lei},
  journal={arXiv preprint arXiv:2412.19442},
  year={2024}
}

@misc{cai2024pyramidkvdynamickvcache,
      title={PyramidKV: Dynamic KV Cache Compression based on Pyramidal Information Funneling}, 
      author={Zefan Cai and Yichi Zhang and Bofei Gao and Yuliang Liu and Tianyu Liu and Keming Lu and Wayne Xiong and Yue Dong and Baobao Chang and Junjie Hu and Wen Xiao},
      year={2024},
      eprint={2406.02069},
      archivePrefix={arXiv},
      primaryClass={cs.CL},
      url={https://arxiv.org/abs/2406.02069}, 
}

@inproceedings{pmlr-v202-liu23am,
  title={Deja vu: Contextual sparsity for efficient llms at inference time},
  author={Liu, Zichang and Wang, Jue and Dao, Tri and Zhou, Tianyi and Yuan, Binhang and Song, Zhao and Shrivastava, Anshumali and Zhang, Ce and Tian, Yuandong and Re, Christopher and others},
  booktitle={International Conference on Machine Learning},
  pages={22137--22176},
  year={2023},
  organization={PMLR}
}

@misc{chen2025powernegativezerodatatype,
      title={RaZeR: Pushing the Limits of NVFP4 Quantization with Redundant Zero Remapping}, 
      author={Yuzong Chen and Xilai Dai and Jake Hyun and Chi-Chih Chang and Wonsuk Jang and Yuheng Wu and Thierry Tambe and Jae-sun Seo and Mohamed S. Abdelfattah},
      year={2026},
      eprint={2501.04052},
      archivePrefix={arXiv},
      primaryClass={cs.LG},
      url={https://arxiv.org/abs/2501.04052}, 
}

@article{akhauri2025tokenbutlertokenimportancepredictable,
      title={TokenButler: Token Importance is Predictable}, 
      author={Yash Akhauri and Ahmed F AbouElhamayed and Yifei Gao and Chi-Chih Chang and Nilesh Jain and Mohamed S. Abdelfattah},
      year={2025},
      eprint={2503.07518},
      archivePrefix={arXiv},
        journal={arXiv preprint arXiv:2503.07518},
      primaryClass={cs.CL},
      url={https://arxiv.org/abs/2503.07518}, 
}

@article{akhauri2024shadowllm,
  title={Shadowllm: Predictor-based contextual sparsity for large language models},
  author={Akhauri, Yash and AbouElhamayed, Ahmed F and Dotzel, Jordan and Zhang, Zhiru and Rush, Alexander M and Huda, Safeen and Abdelfattah, Mohamed S},
  journal={arXiv preprint arXiv:2406.16635},
  year={2024}
}

@misc{dai2024deepseekmoeultimateexpertspecialization,
      title={DeepSeekMoE: Towards Ultimate Expert Specialization in Mixture-of-Experts Language Models}, 
      author={Damai Dai and Chengqi Deng and Chenggang Zhao and R. X. Xu and Huazuo Gao and Deli Chen and Jiashi Li and Wangding Zeng and Xingkai Yu and Y. Wu and Zhenda Xie and Y. K. Li and Panpan Huang and Fuli Luo and Chong Ruan and Zhifang Sui and Wenfeng Liang},
      year={2024},
      eprint={2401.06066},
      archivePrefix={arXiv},
      primaryClass={cs.CL},
      url={https://arxiv.org/abs/2401.06066}, 
}

@misc{perplexityAI,
  title = {Perplexity},
  howpublished = {\url{https://www.perplexity.ai/}},
  note = {Accessed: March 21, 2025},
  author={Perplexity},
  year={2025}
}

@inproceedings{svyatkovskiy2020intellicode,
  title={Intellicode compose: Code generation using transformer},
  author={Svyatkovskiy, Alexey and Deng, Shao Kun and Fu, Shengyu and Sundaresan, Neel},
  booktitle={Proceedings of the 28th ACM joint meeting on European software engineering conference and symposium on the foundations of software engineering},
  pages={1433--1443},
  year={2020}
}

@misc{niah,
  title  = {Needle In A Haystack - Pressure Testing LLMs},
  author = {Kamradt, Greg},
  year   = {2023},
  url    = {https://github.com/gkamradt/LLMTest_NeedleInAHaystack}
}

@article{li2019tutorial_svd,
  title={Tutorial: Complexity analysis of singular value decomposition and its variants},
  author={Li, Xiaocan and Wang, Shuo and Cai, Yinghao},
  journal={arXiv preprint arXiv:1906.12085},
  year={2019}
}

@misc{jiang2024minference10acceleratingprefilling,
      title={MInference 1.0: Accelerating Pre-filling for Long-Context LLMs via Dynamic Sparse Attention}, 
      author={Huiqiang Jiang and Yucheng Li and Chengruidong Zhang and Qianhui Wu and Xufang Luo and Surin Ahn and Zhenhua Han and Amir H. Abdi and Dongsheng Li and Chin-Yew Lin and Yuqing Yang and Lili Qiu},
      year={2024},
      eprint={2407.02490},
      archivePrefix={arXiv},
      primaryClass={cs.CL},
      url={https://arxiv.org/abs/2407.02490}, 
}

@inproceedings{kim2025kvzip,
title={{KV}zip: Query-Agnostic {KV} Cache Compression with Context Reconstruction},
author={Jang-Hyun Kim and Jinuk Kim and Sangwoo Kwon and Jae W. Lee and Sangdoo Yun and Hyun Oh Song},
booktitle={The Thirty-ninth Annual Conference on Neural Information Processing Systems},
year={2025},
url={https://openreview.net/forum?id=JFygzwx8SJ}
}

@misc{yang2025qwen3technicalreport,
      title={Qwen3 Technical Report}, 
      author={An Yang and Anfeng Li and Baosong Yang and Beichen Zhang and Binyuan Hui and Bo Zheng and Bowen Yu and Chang Gao and Chengen Huang and Chenxu Lv and Chujie Zheng and Dayiheng Liu and Fan Zhou and Fei Huang and Feng Hu and Hao Ge and Haoran Wei and Huan Lin and Jialong Tang and Jian Yang and Jianhong Tu and Jianwei Zhang and Jianxin Yang and Jiaxi Yang and Jing Zhou and Jingren Zhou and Junyang Lin and Kai Dang and Keqin Bao and Kexin Yang and Le Yu and Lianghao Deng and Mei Li and Mingfeng Xue and Mingze Li and Pei Zhang and Peng Wang and Qin Zhu and Rui Men and Ruize Gao and Shixuan Liu and Shuang Luo and Tianhao Li and Tianyi Tang and Wenbiao Yin and Xingzhang Ren and Xinyu Wang and Xinyu Zhang and Xuancheng Ren and Yang Fan and Yang Su and Yichang Zhang and Yinger Zhang and Yu Wan and Yuqiong Liu and Zekun Wang and Zeyu Cui and Zhenru Zhang and Zhipeng Zhou and Zihan Qiu},
      year={2025},
      eprint={2505.09388},
      archivePrefix={arXiv},
      primaryClass={cs.CL},
      url={https://arxiv.org/abs/2505.09388}, 
}

@misc{openai2025gptoss120bgptoss20bmodel,
      title={gpt-oss-120b Model Card}, 
      author={OpenAI and : and Sandhini Agarwal and Lama Ahmad and Jason Ai and Sam Altman and Andy Applebaum and Edwin Arbus and Rahul K. Arora and Yu Bai and Bowen Baker and Haiming Bao and Boaz Barak and Ally Bennett and Tyler Bertao and Nivedita Brett and Eugene Brevdo and Greg Brockman and Sebastien Bubeck and Che Chang and Kai Chen and Mark Chen and Enoch Cheung and Aidan Clark and Dan Cook and Marat Dukhan and Casey Dvorak and Kevin Fives and Vlad Fomenko and Timur Garipov and Kristian Georgiev and Mia Glaese and Tarun Gogineni and Adam Goucher and Lukas Gross and Katia Gil Guzman and John Hallman and Jackie Hehir and Johannes Heidecke and Alec Helyar and Haitang Hu and Romain Huet and Jacob Huh and Saachi Jain and Zach Johnson and Chris Koch and Irina Kofman and Dominik Kundel and Jason Kwon and Volodymyr Kyrylov and Elaine Ya Le and Guillaume Leclerc and James Park Lennon and Scott Lessans and Mario Lezcano-Casado and Yuanzhi Li and Zhuohan Li and Ji Lin and Jordan Liss and Lily and Liu and Jiancheng Liu and Kevin Lu and Chris Lu and Zoran Martinovic and Lindsay McCallum and Josh McGrath and Scott McKinney and Aidan McLaughlin and Song Mei and Steve Mostovoy and Tong Mu and Gideon Myles and Alexander Neitz and Alex Nichol and Jakub Pachocki and Alex Paino and Dana Palmie and Ashley Pantuliano and Giambattista Parascandolo and Jongsoo Park and Leher Pathak and Carolina Paz and Ludovic Peran and Dmitry Pimenov and Michelle Pokrass and Elizabeth Proehl and Huida Qiu and Gaby Raila and Filippo Raso and Hongyu Ren and Kimmy Richardson and David Robinson and Bob Rotsted and Hadi Salman and Suvansh Sanjeev and Max Schwarzer and D. Sculley and Harshit Sikchi and Kendal Simon and Karan Singhal and Yang Song and Dane Stuckey and Zhiqing Sun and Philippe Tillet and Sam Toizer and Foivos Tsimpourlas and Nikhil Vyas and Eric Wallace and Xin Wang and Miles Wang and Olivia Watkins and Kevin Weil and Amy Wendling and Kevin Whinnery and Cedric Whitney and Hannah Wong and Lin Yang and Yu Yang and Michihiro Yasunaga and Kristen Ying and Wojciech Zaremba and Wenting Zhan and Cyril Zhang and Brian Zhang and Eddie Zhang and Shengjia Zhao},
      year={2025},
      eprint={2508.10925},
      archivePrefix={arXiv},
      primaryClass={cs.CL},
      url={https://arxiv.org/abs/2508.10925}, 
}

@misc{kimiteam2025kimilinearexpressiveefficient,
      title={Kimi Linear: An Expressive, Efficient Attention Architecture}, 
      author={Kimi Team and Yu Zhang and Zongyu Lin and Xingcheng Yao and Jiaxi Hu and Fanqing Meng and Chengyin Liu and Xin Men and Songlin Yang and Zhiyuan Li and Wentao Li and Enzhe Lu and Weizhou Liu and Yanru Chen and Weixin Xu and Longhui Yu and Yejie Wang and Yu Fan and Longguang Zhong and Enming Yuan and Dehao Zhang and Yizhi Zhang and T. Y. Liu and Haiming Wang and Shengjun Fang and Weiran He and Shaowei Liu and Yiwei Li and Jianlin Su and Jiezhong Qiu and Bo Pang and Junjie Yan and Zhejun Jiang and Weixiao Huang and Bohong Yin and Jiacheng You and Chu Wei and Zhengtao Wang and Chao Hong and Yutian Chen and Guanduo Chen and Yucheng Wang and Huabin Zheng and Feng Wang and Yibo Liu and Mengnan Dong and Zheng Zhang and Siyuan Pan and Wenhao Wu and Yuhao Wu and Longyu Guan and Jiawen Tao and Guohong Fu and Xinran Xu and Yuzhi Wang and Guokun Lai and Yuxin Wu and Xinyu Zhou and Zhilin Yang and Yulun Du},
      year={2025},
      eprint={2510.26692},
      archivePrefix={arXiv},
      primaryClass={cs.CL},
      url={https://arxiv.org/abs/2510.26692}, 
}

@book{Jolliffe2002Principal,
  address = {New York},
  author = {Jolliffe, I. T.},
  booktitle = {Principal Component Analysis},
  comment = {(private-note)Full textbook. I have printed some bits; relevant bit for atm sci is {\S}4.3.},
  doi = {10.1007/b98835},
  isbn = {0-387-95442-2},
  publisher = {Springer-Verlag},
  series = {Springer Series in Statistics},
  title = {Principal Component Analysis},
  url = {http://www.springer.com/statistics/statistical+theory+and+methods/book/978-0-387-95442-4},
  year = 2002
}
\bibliographystyle{icml2026}

\newpage
\appendix
\onecolumn
\section{CKA and Indication of Aligned Left Singular Vectors}
\label{appendix:cka_proof}

\subsection{Notation and Definitions}

For each layer \(\ell\), let
\[
  \mathbf{X}_{\ell} \;\in\; \mathbb{R}^{L\times d},
\]
where each of the \(L\) rows corresponds to a token (data point).  
Define the centering matrix
\[
  \mathbf{H} 
  \;=\; 
  \mathbf{I}_L 
  \;-\; 
  \tfrac{1}{L}\,\mathbf{1}\,\mathbf{1}^\top,
\]
which subtracts the (row) mean from each token embedding.  
We define the \emph{centered} embeddings
\[
  \widetilde{\mathbf{X}}_{\ell}
  \;=\;
  \mathbf{H}\,\mathbf{X}_{\ell},
\]
and the \emph{centered Gram matrix}
\[
  \mathbf{G}_{\ell} 
  \;=\; 
  \widetilde{\mathbf{X}}_{\ell}\,\widetilde{\mathbf{X}}_{\ell}^\top
  \;\in\;
  \mathbb{R}^{L\times L}.
\]
Because \(\mathbf{G}_{\ell}\) is symmetric and positive semidefinite, its largest‐eigenvalue directions capture the most “energetic” dimensions of \(\widetilde{\mathbf{X}}_{\ell}\).

\medskip

Given two layers \(\ell_1\) and \(\ell_2\), the \emph{Centered Kernel Alignment (CKA)} between their token embeddings is
\[
  \mathrm{CKA}\bigl(\mathbf{X}_{\ell_1}, \mathbf{X}_{\ell_2}\bigr)
  \;=\;
  \frac{
    \mathrm{trace}\bigl(\mathbf{G}_{\ell_1}\mathbf{G}_{\ell_2}\bigr)
  }{
    \sqrt{\,\mathrm{trace} \bigl(\mathbf{G}_{\ell_1}^2\bigr)
           \;\mathrm{trace} \bigl(\mathbf{G}_{\ell_2}^2\bigr)\,}
  }
  \,,
\]
which measures how similarly \(\mathbf{G}_{\ell_1}\) and \(\mathbf{G}_{\ell_2}\) encode pairwise relationships (dot products) among the \(n\) token embeddings.

\subsection{SVD Perspective and Overlap}

\paragraph{SVD of centered embeddings.}
Consider the (compact) SVD of \(\widetilde{\mathbf{X}}_{\ell}\):
\[
  \widetilde{\mathbf{X}}_{\ell}
  \;=\;
  \mathbf{U}_{\ell}\,\boldsymbol{\Sigma}_{\ell}\,\mathbf{V}_{\ell}^\top,
\]
where:
\[
  \mathbf{U}_{\ell} \in \mathbb{R}^{L\times r}
  \quad(\text{orthonormal columns}),
  \quad
  \boldsymbol{\Sigma}_{\ell} = \mathrm{diag}(\sigma_{1},\dots,\sigma_{r}),
  \quad
  \mathbf{V}_{\ell} \in \mathbb{R}^{d\times r}
  \quad(\text{orthonormal columns}),
\]
and \(r \le \min(n, d)\) is the rank.  Then the centered Gram matrix factors as
\[
  \mathbf{G}_{\ell}
  \;=\;
  \widetilde{\mathbf{X}}_{\ell}\,\widetilde{\mathbf{X}}_{\ell}^\top
  \;=\;
  \mathbf{U}_{\ell}
  \,\boldsymbol{\Sigma}_{\ell}^2
  \,\mathbf{U}_{\ell}^\top,
\]
so the columns of \(\mathbf{U}_{\ell}\) are exactly the eigenvectors of \(\mathbf{G}_{\ell}\), and \(\sigma_{i}^2\) are the corresponding eigenvalues.

\paragraph{CKA in terms of left singular vectors.}
Let \(\widetilde{\mathbf{X}}_{\ell_1} = \mathbf{U}_{\ell_1}\,\boldsymbol{\Sigma}_{\ell_1}\,\mathbf{V}_{\ell_1}^\top\) and 
\(\widetilde{\mathbf{X}}_{\ell_2} = \mathbf{U}_{\ell_2}\,\boldsymbol{\Sigma}_{\ell_2}\,\mathbf{V}_{\ell_2}^\top\).  
Then
\[
  \mathbf{G}_{\ell_1}
  \;=\;
  \mathbf{U}_{\ell_1}\,\boldsymbol{\Sigma}_{\ell_1}^2\,\mathbf{U}_{\ell_1}^\top,
  \quad
  \mathbf{G}_{\ell_2}
  \;=\;
  \mathbf{U}_{\ell_2}\,\boldsymbol{\Sigma}_{\ell_2}^2\,\mathbf{U}_{\ell_2}^\top.
\]
We compute 
\[
  \mathrm{trace}\!\bigl(\mathbf{G}_{\ell_1}\,\mathbf{G}_{\ell_2}\bigr)
  \;=\;
  \mathrm{trace}\!\Bigl(
    \mathbf{U}_{\ell_1}\,\boldsymbol{\Sigma}_{\ell_1}^2\,\mathbf{U}_{\ell_1}^\top
    \;\mathbf{U}_{\ell_2}\,\boldsymbol{\Sigma}_{\ell_2}^2\,\mathbf{U}_{\ell_2}^\top
  \Bigr)
  \;=\;
  \sum_{i=1}^{r_1} \sum_{j=1}^{r_2}
    \sigma_{\ell_1,i}^2\,\sigma_{\ell_2,j}^2
    \,\Bigl(\mathbf{u}_{\ell_1}^{(i)}{}^\top \mathbf{u}_{\ell_2}^{(j)}\Bigr)^2,
\]
where \(\mathbf{u}_{\ell_1}^{(i)}\) and \(\mathbf{u}_{\ell_2}^{(j)}\) are the \(i\)-th and \(j\)-th columns of \(\mathbf{U}_{\ell_1}\) and \(\mathbf{U}_{\ell_2}\), respectively.  Meanwhile,
\[
  \mathrm{trace}\!\bigl(\mathbf{G}_{\ell_1}^2\bigr)
  \;=\;
  \sum_{i=1}^{r_1}
    \sigma_{\ell_1,i}^4,
  \qquad
  \mathrm{trace}\!\bigl(\mathbf{G}_{\ell_2}^2\bigr)
  \;=\;
  \sum_{j=1}^{r_2}
    \sigma_{\ell_2,j}^4.
\]
Hence, 
\[
  \mathrm{CKA}\bigl(\mathbf{X}_{\ell_1}, \mathbf{X}_{\ell_2}\bigr)
  \;=\;
  \frac{
    \displaystyle \sum_{i,j}
      \sigma_{\ell_1,i}^2\,\sigma_{\ell_2,j}^2
      \bigl(\mathbf{u}_{\ell_1}^{(i)}{}^\top \mathbf{u}_{\ell_2}^{(j)}\bigr)^2
  }{
    \sqrt{\Bigl(\sum_{i}\sigma_{\ell_1,i}^4\Bigr)
          \Bigl(\sum_{j}\sigma_{\ell_2,j}^4\Bigr)}
  }
  \,.
\]
Because the eigenvalues \(\sigma_{\ell,i}^2\) reflect how “dominant” each left singular vector is, a \textbf{large CKA} value requires significant overlap 
\(\bigl(\mathbf{u}_{\ell_1}^{(i)}{}^\top \mathbf{u}_{\ell_2}^{(j)}\bigr)^2\) for the most important (largest‐\(\sigma^2\)) directions, implying the principal subspaces of \(\mathbf{G}_{\ell_1}\) and \(\mathbf{G}_{\ell_2}\) align closely.

\subsection{Conclusion}

In summary, when \(\mathrm{CKA}(\mathbf{X}_{\ell_1}, \mathbf{X}_{\ell_2})\) is high, the dominant \emph{left singular vectors} of \(\widetilde{\mathbf{X}}_{\ell_1}\) and \(\widetilde{\mathbf{X}}_{\ell_2}\) are well aligned.  
Since these vectors also serve as the \emph{largest‐eigenvalue} directions of the centered Gram matrices, high CKA implies that the \emph{principal subspace} geometry of the token embeddings in layers~\(\ell_1\) and~\(\ell_2\) is \emph{structurally} very similar—even if token‐by‐token (cosine) matches are small.  
Thus, CKA goes beyond individual token similarities, capturing \textbf{how} tokens vary collectively in a shared subspace.

\section{Implementation Details}
\subsection{Landmark-guided Chunk Selector for Selective Reconstruction}
\label{appendix: landmark detail}


\begin{algorithm}[ht]
\caption{Landmark Construction (Prefill)}
\label{alg:landmark_construction_tensor}
\begin{algorithmic}[1]
\REQUIRE Post-RoPE keys $\mathbf{K}^{\mathrm{rope}}_{\ell}\!\in\!\mathbb{R}^{H_{\mathrm{kv}}\times L\times d_h}$, chunk size $c$, optional \#outliers $o$
\ENSURE Landmarks $\mathbf{L}_{\ell}\!\in\!\mathbb{R}^{H_{\mathrm{kv}}\times n_c\times d_h}$, optional outlier indices $\{\mathcal{O}_{\ell,g}\!\subseteq\![n_c]\}_{g=1}^{H_{\mathrm{kv}}}$
\STATE $n_c \gets \lceil L/c\rceil$
\STATE \textbf{Chunking the sequence:} $\widetilde{\mathbf{K}}\gets \mathrm{View}\!\left(\mathbf{K}^{\mathrm{rope}}_{\ell}\right)\in\mathbb{R}^{H_{\mathrm{kv}}\times n_c\times c\times d_h}$
\STATE \textbf{Chunk means (landmarks):} $\mathbf{L}_{\ell}\;\gets\;\mathrm{mean}\!\left(\widetilde{\mathbf{K}},\,\text{axis}=2\right)\in\mathbb{R}^{H_{\mathrm{kv}}\times n_c\times d_h}$
\STATE \textbf{(Optional) Static outliers, per head:} $\mathbf{S}_{\cos}\!\gets\!\mathrm{cos}\!\big(\widetilde{\mathbf{K}},\,\mathbf{L}_{\ell}\text{ broadcast along }c\big)\in\mathbb{R}^{H_{\mathrm{kv}}\times n_c\times c}$
\STATE \hspace{2em}$m\!\gets\!\min(\mathbf{S}_{\cos},\text{axis}=2)\in\mathbb{R}^{H_{\mathrm{kv}}\times n_c}$;\quad $\mathbf{I}^{\text{out}}\!\gets\!\mathrm{ArgTopK}(-m,\,o)$;\quad $\mathcal{O}_{\ell,g}\!\gets\!\mathbf{I}^{\text{out}}[g,:]$
\STATE \textbf{return} $\mathbf{L}_{\ell}$ and (optionally) $\{\mathcal{O}_{\ell,g}\}$
\end{algorithmic}
\end{algorithm}

\begin{algorithm}[ht]
\caption{Landmark-Guided Top-$k$ Chunk Selection (Decode)}
\label{alg:selector_token_budget_tensor}
\begin{algorithmic}[1]
\REQUIRE Landmarks $\mathbf{L}_{\ell}\!\in\!\mathbb{R}^{H_{\mathrm{kv}}\times n_c\times d_h}$, queries $\mathbf{Q}_{t,\ell}\!\in\!\mathbb{R}^{H_q\times d_h}$,
         GQA map $\rho:[H_q]\!\to\![H_{\mathrm{kv}}]$, token budget $k$, chunk size $c$,
         optional outliers $\{\mathcal{O}_{\ell,g}\}$
\ENSURE Per–KV head selected chunk indices $\{S_{t,\ell,g}\!\subseteq\![n_c]\}_{g=1}^{H_{\mathrm{kv}}}$
\STATE $k_{\mathrm{ch}} \gets \left\lceil k / c \right\rceil$ \hfill \textit{// convert token budget to chunk budget}
\STATE \textbf{Scores to landmarks (batched MatMul):}
       \[
         \mathbf{P} \in \mathbb{R}^{H_q\times H_{\mathrm{kv}}\times n_c}
         \;\gets\; \left\langle \mathbf{Q}_{t,\ell}[:,\cdot],\, \mathbf{L}_{\ell}[\cdot,:, \cdot]\right\rangle_{d_h}\big/\sqrt{d_h}
       \]
\STATE \textbf{Pool from query heads to KV heads (GQA):}
       \[
         \mathbf{S}[g,j] \;\gets\; \max_{\,h:\,\rho(h)=g}\; \mathbf{P}[h,g,j]
         \quad\text{for all } g\in[H_{\mathrm{kv}}],~j\in[n_c]
       \]
\STATE \textbf{Top-$k_{\mathrm{ch}}$ per KV head:}\quad
       $\mathbf{I} \in \mathbb{R}^{H_{\mathrm{kv}}\times k_{\mathrm{ch}}}\;\gets\;\mathrm{ArgTopK}\!\left(\mathbf{S},\,k_{\mathrm{ch}}\right)$
\STATE \textbf{(Optional) add static outliers:}\quad
       $S_{t,\ell,g}\;\gets\;\mathrm{Union}\!\big(\mathbf{I}[g,:],\,\mathcal{O}_{\ell,g}\big)$ \quad for each $g$
\STATE \textbf{return} $\{S_{t,\ell,g}\}_{g=1}^{H_{\mathrm{kv}}}$
\end{algorithmic}
\end{algorithm}

While our core compression method treats keys and values uniformly, the token selection mechanism operates exclusively on the \textbf{Key cache} to estimate attention importance.
In this section, we denote the full \textbf{post-RoPE key cache} at layer $\ell$ as $\mathbf{K}^{\mathrm{rope}}_{\ell} \in \mathbb{R}^{H_{\mathrm{kv}} \times L \times d_h}$, where $L$ is the sequence length. We adopt the landmark-guided selection technique from ShadowKV \citep{sun2024shadowkvkvcacheshadows}.

\paragraph{Landmark construction (prefill).}
At layer $\ell$, we split the post-RoPE key sequence into $n_c=\lceil L/c\rceil$ contiguous chunks of size $c$. For each KV head $g$ and chunk $j$, we define the \emph{landmark} as the mean key of that chunk:
\[
  \mathbf{\ell}_{j,g} \;=\; \frac{1}{|C_j|}\sum_{x\in C_j} \mathbf{K}^{\mathrm{rope}}_{\ell,g}(x).
\]
Optionally, we maintain a small set of \emph{static outliers} per head to guard against heterogeneous chunks where the mean is a poor representative. We identify these by computing the minimum within-chunk cosine similarity to the landmark,
\[
  r_{g,j} \;=\; \min_{x\in C_j}\, \cos\!\big(\mathbf{K}^{\mathrm{rope}}_{\ell,g}(x),\, \mathbf{\ell}_{j,g}\big),
\]
and marking the $o$ chunks with the smallest $r_{g,j}$ as outliers. Lower values indicate that the chunk contains tokens significantly distinct from the mean; thus, these chunks are always preserved during decoding. The procedure is summarized in Algorithm~\ref{alg:landmark_construction_tensor}.

\paragraph{Landmark-guided selection (decode).}
At each decode step $t$, given queries $\mathbf{Q}_{t,\ell}\!\in\!\mathbb{R}^{H_q\times d_h}$, we score every chunk via a batched scaled dot-product between $\mathbf{Q}_{t,\ell}$ and the landmarks. Under Grouped-Query Attention (GQA), scores are pooled from query heads to KV heads using the mapping $\rho:[H_q]\!\to\![H_{\mathrm{kv}}]$ by taking the maximum score over the query heads mapped to each KV head.
Given a token budget $k$, we define a chunk budget $k_{\mathrm{ch}}=\lceil k/c\rceil$ and select the top $k_{\mathrm{ch}}$ chunks per KV head. Optionally, we union these with the static outliers $\mathcal{O}_{\ell,g}$. The selected chunk indices are then expanded to row indices $\mathcal{S}_{t,\ell,g}$ for Selective Reconstruction. This workflow is detailed in Algorithm~\ref{alg:selector_token_budget_tensor}.

\section{More latency studies}
\subsection{On-the-fly SVD overhead}
\label{App: Prefill SVD Time}
Table~\ref{tab:svd_latency_cholesky_llama_a100} reports the latency of the prefilling phase as well as the cross-layer SVD using our custom kernel on an A6000 GPU. On a sequence length of $L=64\text{k}$ tokens (with $W=4$), the SVD overhead accounts for only 3.17\% of the forward-pass time. This fraction steadily decreases as $L$ increases, dropping to a mere 0.84\% at $L=256\text{k}$. This reduction can be attributed to the fact that the cost of attention grows quadratically with $L$, whereas our low-rank decomposition scales only linearly~\citep{li2019tutorial_svd}. As a result, for very long contexts, the one-time decomposition performed during the prefill phase becomes practically negligible, contributing minimally to the overall computation time. Similar scaling trends also hold across different model architectures, such as Qwen2.5-14B, as demonstrated in Table~\ref{tab:svd_latency_cholesky_qwen_a100}.



\begin{table}[!h]
    \centering
    \caption{Latency data for on-the-fly SVD across different context lengths. Measured on an A100 GPU with Llama-3.1-8B. (Unit: seconds)}
    \vspace{5pt}
    \begin{tabular}{l|cccc}
        \toprule
        Seqlen & 64k & 128k & 160k & 256k \\
        \midrule
        Prefill Time & 9.97 & 32.00 & 47.84 & 113.67 \\
        SVD time ($W$=2) & 0.42 (4.21\%) & 0.58 (1.81\%) & 0.67 (1.40\%) & 0.99 (0.87\%) \\
        SVD time ($W$=4) & 0.57 (5.72\%) & 0.80 (2.50\%) & 0.92 (1.92\%) & 1.41 (1.24\%) \\
        \bottomrule
    \end{tabular}
    \label{tab:svd_latency_cholesky_llama_a100}
\end{table}

\begin{table}[!h]
    \centering
    \caption{Latency data for on-the-fly SVD across different context lengths. Measured on an A100 GPU with Qwen2.5-14B-Instruct. (Unit: seconds)}
    \vspace{5pt}
    \begin{tabular}{l|cccc}
        \toprule
        Seqlen & 64k & 128k & 160k & 256k \\
        \midrule
        Prefill Time & 18.60 & 60.03 & 89.76 & 212.99 \\
        SVD time ($W$=2) & 0.62 (3.35\%) & 0.87 (1.45\%) & 1.00 (1.11\%) & 1.48 (0.69\%) \\
        SVD time ($W$=4) & 0.85 (4.55\%) & 1.20 (2.00\%) & 1.38 (1.54\%) & 2.11 (0.99\%) \\
        \bottomrule
    \end{tabular}
    \label{tab:svd_latency_cholesky_qwen_a100}
\end{table}

\subsection{Custom SVD Kernel}
\label{App: Custom SVD Kernel}
To accelerate online \kv{} compression during the prefill phase, we developed a custom randomized SVD kernel that overcomes the memory-bandwidth bottlenecks of standard implementations like \texttt{torch.svd\_lowrank}. Instead of relying on the sequential FP32 Householder QR, our kernel introduces two hardware-aware optimizations: executing power iterations with 16-bit matrix multiplications to exploit GPU Tensor Cores, and replacing Householder QR with a parallelizable, shifted Cholesky QR factorization. This approach successfully transforms memory-bound orthogonalization into compute-bound GEMMs while maintaining strict numerical stability in lower precision. Ultimately, our kernel delivers a $4\times$ speedup over the standard PyTorch API, reducing decomposition overhead to a negligible fraction without compromising model accuracy.

\section{More Experimental Results}
\label{app: More Experimental Results}

\subsection{More Results on RULER}
\label{app: More Results on RULER}

\paragraph{\kv{} More Compression with Selective Reconstruction Results.}
Table~\ref{tab:ruler_sparse_full} reports results at different compression rates on the RULER benchmark. 
At the high compression setting (around $11.5\times$ effective GPU memory reduction), \xKSR{} outperforms ShadowKV by a striking 36\%. This demonstrates that \xKVSR{} is significantly more effective at preserving performance under extreme compression.

\begin{table}[h]
    \centering
    \caption{More \kv{} Compression with Selective Reconstruction Results: Accuracy of different methods on the RULER benchmark at a context length of 64K. Here, "Comp." indicates the total KV-Cache reduction, while the number in parentheses shows the effective GPU memory reduction considering \kv{} offloading. ShadowKV* refers to a variant of ShadowKV that additionally compresses the value cache.}
    \vspace{-5pt}
    \resizebox{48em}{!}{
    \begin{tabular}{r|c|cccccccccc|c}
    \toprule
    Method & Comp. & N-S1 & N-S2 & N-MK1 & N-MK2 & N-MQ & N-MV & QA-1 & QA-2 & VT & FWE & Avg. \\
    \midrule
    \multicolumn{13}{c}{\textbf{Llama-3.1-8B-Instruct}} \\
    \midrule

    Full Attn   & 1.00 & 100.00 & 100.00 & 98.96 & 97.92 & 98.96 & 97.66 & 83.33 & 59.38 & 97.29 & 85.42 & 91.89 \\
    Quest      & 1.00 (8.00) & 93.75  & 90.63  & 96.88 & 87.50 & 94.27 & 85.42 & 83.33 & 57.29 & 77.71 & 81.94 & 84.87 \\

    \midrule
    ShadowKV & 1.60 (7.94) & 100.00 & 100.00 & 100.00 & 97.92 & 99.22 & 95.83 & 83.33 & 59.38 & 78.33 & 73.96 & 88.80 \\
    \rowcolor{lightmintbg}
    \xKSR{} (Ours)  & 1.59 (7.76) & 100.00 & 100.00 & 98.96 & 97.92 & 98.70 & 96.35 & 82.29 & 61.46 & 88.33 & 75.69 & 89.97 \\
    ShadowKV & 1.64 (9.08) & 100.00 & 100.00 & 98.96 & 97.92 & 96.88 & 94.53 & 82.29 & 60.42 & 66.04 & 74.65 & 87.17 \\
    \rowcolor{lightmintbg}
    \xKSR{} (Ours)  & 1.63 (8.90) & 100.00 & 100.00 & 98.96 & 97.92 & 98.44 & 95.31 & 83.33 & 60.42 & 87.92 & 74.65 & 89.70 \\
    ShadowKV   & 1.68 (10.61) & 100.00 & 71.88  & 73.96 & 97.92 & 27.34 & 24.22 & 68.75 & 58.33 & 52.71 & 73.96 & 64.91 \\
    \rowcolor{lightmintbg}
    \xKSR{} (Ours)  & 1.68 (10.45)& 100.00 & 98.96  & 98.96 & 97.92 & 94.53 & 93.49 & 82.29 & 60.42 & 80.83 & 76.04 & 88.34 \\
    ShadowKV & 1.71 (11.59) & 96.88 & 6.25 & 5.21 & 80.21 & 0.78 & 2.34 & 65.62 & 56.25 & 49.79 & 72.57 & 43.59 \\
    \rowcolor{lightmintbg}
    \xKSR{} (Ours)  & 1.70 (11.44) & 100.00 & 96.88 & 92.71 & 97.92 & 62.50 & 56.25 & 80.21 & 59.38 & 69.58 & 76.39 & 79.18 \\

    \midrule
    ShadowKV* & 4.52 & 100.00 & 98.96 & 96.88 & 97.92 & 93.49 & 91.67 & 82.29 & 58.33 & 67.92 & 75.69 & 86.32 \\
    \rowcolor{lightmintbg}
    \xKVSR{} (Ours) & 4.37 & 100.00 & 100.00 & 98.96 & 96.88 & 99.48 & 96.61 & 82.29 & 60.42 & 87.92 & 75.69 & 89.83 \\
    ShadowKV*   & 5.51 & 100.00 & 76.04 & 75.00 & 97.92 & 54.43 & 45.83 & 81.25 & 57.29 & 47.29 & 74.31 & 70.94 \\
    \rowcolor{lightmintbg}
    \xKVSR{} (Ours)  & 5.35 & 100.00 & 100.00 & 98.96 & 97.92 & 98.44 & 95.57 & 82.29 & 60.42 & 87.29 & 76.04 & 89.69 \\

    \bottomrule
    \end{tabular}
    }
    \label{tab:ruler_sparse_full}
\end{table}

\subsection{Results on LongBench}
\label{app: Results on LongBench}
\paragraph{\kv{} Compression Results.}
Table \ref{tab:longbench_comp} presents the comprehensive evaluation of \xKV{} against representative compression methods on the LongBench dataset, demonstrating consistent performance across diverse long-context tasks, including single-document QA, multi-document QA, summarization, few-shot learning, synthetic tasks, and code completion. Experiments were conducted on Llama-3.1-8B-Instruct and Qwen2.5-7B-Instruct-1M models.

MiniCache exhibits severe performance degradation, with accuracy dropping by 12.57\% on Llama-3.1-8B-Instruct and a catastrophic 26.91\% on Qwen2.5-7B-Instruct-1M compared to the baseline, reinforcing our earlier observation that cross-layer compression methods fail when token-wise cosine similarity assumptions are violated across different model architectures and task types.

At 8.03× compression, \xKV{} achieves 42.27\% average accuracy on Llama-3.1-8B-Instruct, closely matching PyramidKV and SnapKV. On Qwen2.5-7B-Instruct-1M, \xKV{} reaches 40.37\% accuracy, demonstrating competitive performance against PyramidKV and SnapKV, with a slight accuracy degradation.

These LongBench results validate \xKV{}'s robustness across heterogeneous task domains, confirming that our shared low-rank subspace approach effectively preserves critical information for diverse long-context reasoning scenarios while achieving aggressive compression rates comparable to leading token eviction methods.

\begin{table}[!t]
    \centering
    \caption{\kv{} Compression Results: Accuracy of different methods on LongBench. \xKV{} consistently achieves a higher accuracy than the Full Attns at the same compression rate or even at a significantly higher compression rate.}
    \vspace{-5pt}
    \resizebox{48em}{!}{
    \begin{tabular}{r|c|c|cccccc|c}
    \toprule
    Method & Type & Comp. & Single-doc QA & Multi-doc QA & Summarization & Few-shot & Synthetic & Code & Avg. \\
    \midrule
    \multicolumn{9}{c}{\textbf{Llama-3.1-8B-Instruct}} \\
    \midrule
    Full Attn     &     - & 1.00 & 44.23 & 44.72 & 28.52 & 25.88 & 53.44 & 62.41 & 43.20 \\
    \midrule
    KIVI-2        & Intra & 7.10 & 40.87 & 42.45 & 27.40 & 26.96 & 51.70 & 59.42 & 41.47 \\
    StreamingLLM  & Intra & 8.00 & 30.04 & 37.79 & 23.61 & 25.49 & 49.75 & 61.15 & 37.97 \\
    PyramidKV     & Intra & 8.00 & 42.92 & 43.99 & 25.73 & 27.62 & 53.02 & 61.54 & 42.47 \\
    SnapKV        & Intra & 8.00 & 43.17 & 44.13 & 26.09 & 27.75 & 53.27 & 62.56 & 42.83 \\
    Single SVD    & Intra & 8.40 & 30.34 & 23.93 & 20.26 & 27.41 & 44.75 & 52.63 & 33.22 \\
    \midrule
    MiniCache     & Inter & 1.30 & 22.01 & 26.79 & 20.51 & 25.05 & 52.29 & 37.11 & 30.63 \\
    \rowcolor{lightmintbg}
    \xKV{} (Ours) & Inter & 8.03 & 44.39 & 38.82 & 26.14 & 27.34 & 55.50 & 61.44 & 42.27 \\
    \bottomrule
    \end{tabular}
    }
    \label{tab:longbench_comp}
\end{table}

\paragraph{\kv{} Compression with Selective Reconstruction Results.}
In Table~\ref{tab:longbench_sparse}, we evaluate \xKSR{} and \xKVSR{} against Quest and ShadowKV baselines on the LongBench dataset using Llama-3.1-8B-Instruct. Quest achieves 42.63\% accuracy through dynamic token loading with 8× GPU memory reduction via offloading, demonstrating minimal performance degradation while requiring host-device transfers.

At comparable compression ratios, \xKSR{} consistently outperforms ShadowKV across different settings. With 1.68× compression and 10.45× GPU memory reduction, \xKSR{} achieves 42.50\% accuracy, surpassing ShadowKV by 1.99\%. This improvement demonstrates the effectiveness of our cross-layer key compression approach over single-layer SVD methods.

Most notably, \xKVSR{} enables aggressive 5.35× compression while achieving 42.40\% accuracy, outperforming ShadowKV* by 0.89\%. These consistent improvements across both RULER and LongBench benchmarks validate that our cross-layer alignment approach effectively adapts to diverse evaluation frameworks, preserving critical information across heterogeneous long-context tasks ranging from retrieval and reasoning to code completion and summarization. Moreover, the significant gains observed on LongBench further corroborate the robustness and generality of our method beyond the RULER benchmark.

\begin{table}[!t]
    \centering
    \caption{\kv{} Compression with Selective Reconstruction Results: Accuracy of different methods on the LongBench. Here, "Comp." indicates the total memory reduction, while the number in parentheses shows the effective GPU memory reduction considering \kv{} offloading. ShadowKV* refers to a variant of ShadowKV that additionally compresses the value cache.}
    \vspace{-5pt}
    \resizebox{48em}{!}{
    \begin{tabular}{r|c|cccccc|c}
    \toprule
    Method & Comp. & Single-doc QA & Multi-doc QA & Summarization & Few-shot & Synthetic & Code & Avg. \\
    \midrule
    \multicolumn{9}{c}{\textbf{Llama-3.1-8B-Instruct}} \\
    \midrule
    Full Attn  & 1.00         & 44.23 & 44.72 & 28.52 & 25.88 & 53.44 & 62.41 & 43.20 \\
    Quest     & 1.00 (8.00)  & 43.18 & 44.40 & 28.20 & 26.57 & 52.88 & 60.55 & 42.63 \\

    \midrule
    ShadowKV  & 1.68 (10.61) & 37.98 & 44.11 & 25.26 & 24.43 & 53.35 & 57.92 & 40.51 \\
    \rowcolor{lightmintbg}
    \xKSR{}   & 1.68 (10.45) & 43.64 & 44.47 & 27.62 & 25.31 & 52.63 & 61.32 & 42.50 \\

    ShadowKV & 1.64 (9.08) & 43.35 & 44.87 & 27.15 & 25.76 & 52.63 & 59.53 & 42.21 \\
    \rowcolor{lightmintbg}
    \xKVSR{} & 1.63 (8.90) & 44.38 & 44.63 & 27.98 & 25.55 & 52.13 & 61.50 & 42.69 \\
    
    
    \midrule
    ShadowKV* & 5.51 & 41.76 & 44.89 & 26.02 & 24.74 & 52.73 & 58.91 & 41.51 \\
    \rowcolor{lightmintbg}
    \xKVSR{}  & 5.35 & 44.58 & 45.20 & 27.76 & 25.32 & 52.63 & 58.94 & 42.40 \\
    \bottomrule
    \end{tabular}
    }
    \label{tab:longbench_sparse}
\end{table}



\begin{figure}[ht]
    \centering
    \includegraphics[width=1.0\linewidth]{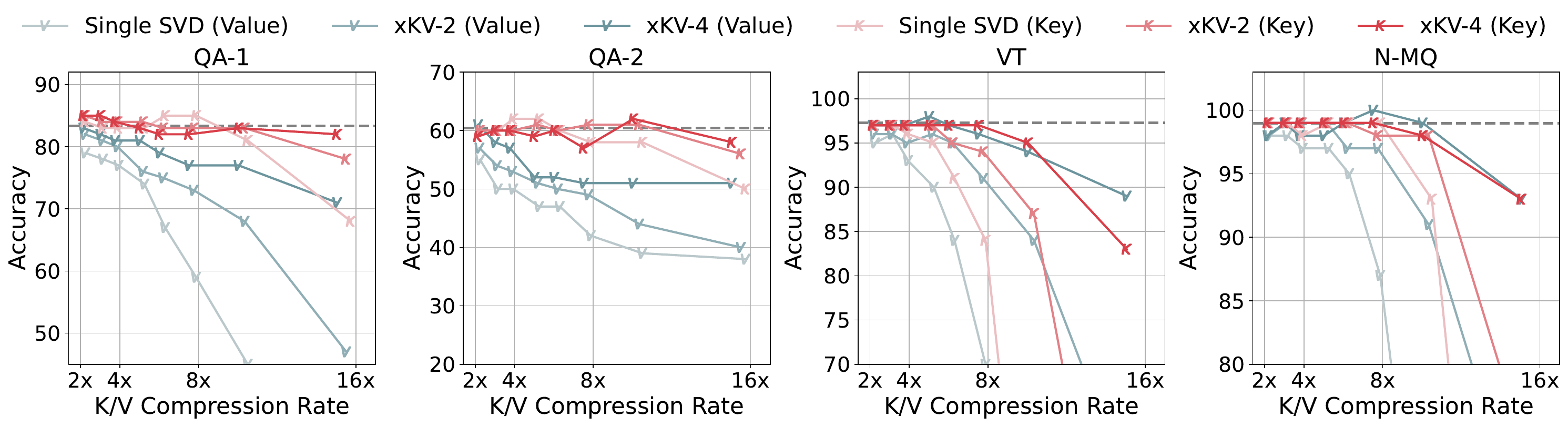}
    \caption{Accuracy comparison of applying different methods to key and value separately on Llama-3.1-8B-Instruct using RULER benchmark. The number after \xKV{} denotes the cross-layer window size.}
    \label{fig:kv_ablation}
\end{figure}

\subsection{Impact of \xKV{} on Compressing Value and Key Only.}
To understand how \xKV{} affects key and value compression, we conduct ablation experiments on four subtasks from RULER \citep{hsieh2024ruler} to evaluate how \xKV{} (cross-layer low-rank SVD) affects key and value compression. We show the results in Figure~\ref{fig:kv_ablation}. Overall, \xKV{} consistently boosts accuracy under varying compression rates. Also, keys exhibit higher compressibility than values, matching the eigenvalue analysis in Figure~\ref{fig:sub3}. A closer inspection of the results reveals that the achievable compression ratio appears to be task-dependent. On the questions-answering subtasks (QA-1 and QA-2) \xKV{} can push the compression rate to 16$\times$ while still preserving performance. In Variable Tracking (VT) and NIAH multi-queries (N-MQ) \citep{niah}, accuracy begins to decline beyond 8$\times$ compression; however, in these same tasks, values tolerate compression more easily than in QA subtasks. 
These observations underscore how different tasks may demand different “sweet spots” for key versus value compression. In \xKV{}, we employ a fixed compression ratio for all different tasks. Exploring task-specific or context-aware \citep{pmlr-v202-liu23am, akhauri2025tokenbutlertokenimportancepredictable, akhauri2024shadowllm} rank allocation is a promising avenue for future work.

\subsection{Impact of Cross-layer Window Size on Accuracy.}
\begin{table}[htp]
  \centering
  \captionof{table}{Accuracy across different window sizes on RULER with Llama-3.1-8B-Instruct. 
  We align the rank setting with Table~\ref{tab:ruler_comp} and Table~\ref{tab:ruler_sparse} for window size 4. For window sizes 1, 2, and 8, we scaled the rank linearly to maintain the same compression rate, with \((r_{K^{pre}},r_V)=(96,144)\) and \((192,288)\), respectively.}
  \label{tab:xkv_gs_study}
  \resizebox{14em}{!}{
  \begin{tabular}{c|c|c|c}
      \toprule
      \makecell{Window \\ Size} & \makecell{\xKV{}} & \makecell{\xKSR{}} & \makecell{\xKVSR{}} \\
      \midrule
      1 & 45.71 & 87.17 & 72.27 \\
      2 & 75.15 & 88.43 & 86.06 \\
      4 & 88.50 & 89.70 & 89.69 \\
      8 & 88.91 & 89.74 & 89.72 \\
      \bottomrule
  \end{tabular}
  }
\end{table}

\label{subec: Window Size Study}
To quantify the impact of cross-layer compression, we conduct a window size ablation on the RULER benchmark at a fixed compression rate (Table~\ref{tab:xkv_gs_study}). For example, \xKV{} improves from 45.71\% with window size 1 to 75.15\% at size 2, and further to 88.50\% at size 4. Similar trends are observed for \xKSR{} and \xKVSR{}, where performance likewise climbs steadily as window size increases. These results confirm that sharing across more layers consistently enhances reconstruction fidelity under an identical compression rate.
However, at a window size of 8, the accuracy of \xKV{}, \xKSR{}, and \xKVSR{} all saturates, with accuracy nearly identical to that at a size of 4. Therefore, we use a window size of 4 in all main experiments as it maximizes accuracy while keeping the prefill buffering overhead minimal compared to larger window sizes.

\subsection{Integrate with Quantization}
\label{app: Integrate with Quantization}
\begin{table}[!h]
    \centering
    \caption{Accuracy of \xKV{} integrated with naive quantization on RULER benchmark.}
    \vspace{-5pt}
    \resizebox{48em}{!}{
    \begin{tabular}{r|c|ccccccccccc}
        \toprule
        Method & Comp. & N-S1 & N-S2 & N-MK1 & N-MK2 & N-MQ & N-MV & QA-1 & QA-2 & VT & FWE & Avg. \\
        \midrule
        \multicolumn{13}{c}{\textbf{Llama-3.1-8B-Instruct}} \\
        \midrule
        Full Attn     & 1.00  & 100.00 & 100.00 & 98.96 & 97.92 & 98.96 & 98.18 & 83.33 & 60.42 & 97.71 & 85.42 & 92.09 \\
        \xKV{}       & 8.03  & 100.00 & 98.96  & 97.92 & 97.92 & 96.35 & 97.14 & 78.13 & 57.29 & 86.67 & 78.13 & 88.85 \\
        \xKV{}-4bit  & 25.70 & 100.00 & 96.88  & 97.92 & 97.92 & 96.35 & 93.23 & 77.08 & 55.21 & 83.33 & 78.47 & 87.64 \\
        \xKV{}-3bit  & 32.12 & 93.75  & 94.79  & 95.83 & 96.88 & 95.05 & 90.89 & 77.08 & 52.08 & 73.33 & 76.74 & 84.64 \\
        \bottomrule
    \end{tabular}
    }
    \label{tab:ruler_quant}
\end{table}

\xKV{} can be combined with other cache management techniques. To illustrate this capability, we conducted preliminary experiments integrating \xKV{} with Quantization. Specifically, we applied a simple round-to-nearest (RTN) quantization method to the compressed cache. With 4-bit quantization, the cache achieves a substantial 25.6$\times$ compression while maintaining model accuracy.

Table~\ref{tab:ruler_quant} presents the performance of \xKV{} with naive quantization on the RULER benchmark, evaluated using Llama-3.1-8B-Instruct. We observe that \xKV{} alone provides an 8$\times$ compression with minimal accuracy loss. Further applying 4-bit quantization yields a total compression of 25.6$\times$, with only a slight drop in the average score from 88.85\% to 87.64\%. Even more aggressive 3-bit quantization achieves 32$\times$ compression, with a moderate decrease in performance (average 84.64\%), demonstrating that \xKV{} can be effectively combined with other cache reduction techniques without severely impacting accuracy.

\subsection{FLOPs \& Memory Cost}
\label{app:flops-memory}

\paragraph{Setup and notation.}
We consider a Transformer with $N$ decoder layers and a prompt of length $L$.
Under GQA, there are $H_{\mathrm{kv}}$ KV heads, each with per-head width $d_h$,
so the full KV hidden size is $d := H_{\mathrm{kv}} d_h$.
As in \S\ref{sec:cls}, layers are partitioned into windows of size $W$; for any layer $\ell$,
let $k := \lfloor \ell/W \rfloor$ denote the window index that contains layer $\ell$.
Cross-Layer Factorization (CLF) is applied \emph{independently} to the \textbf{pre-RoPE key cache} and the \textbf{value cache}.
We allow different ranks $r_K$ (keys) and $r_V$ (values).

\vspace{2pt}
\noindent
\textbf{Key factors:} $\mathbf{A}^{K}_k \in \mathbb{R}^{L\times r_K}$ and $\mathbf{B}^{K}_{\ell}\in\mathbb{R}^{r_K\times d}$
(with head blocks $\mathbf{B}^{K}_{\ell,g}\in\mathbb{R}^{r_K\times d_h}$).
\textbf{Value factors:} $\mathbf{A}^{V}_k \in \mathbb{R}^{L\times r_V}$ and $\mathbf{B}^{V}_{\ell}\in\mathbb{R}^{r_V\times d}$
(with head blocks $\mathbf{B}^{V}_{\ell,g}\in\mathbb{R}^{r_V\times d_h}$).
(If only keys are compressed, the value terms are omitted.)

\paragraph{Dense reconstruction (no selection).}
If we reconstruct all $L$ rows per layer at decode, then for one layer $\ell$ the dense reconstruction cost is:
\begin{equation}
\label{eq:cost-dense}
\mathrm{FLOPs}_{\mathrm{dense}}
\;=\;
\underbrace{L\,r_K\,d}_{\text{keys}}
\;+\;
\underbrace{L\,r_V\,d}_{\text{values (if compressed)}}.
\end{equation}
(We ignore constant factors such as the multiply-add factor of $2$; this does not affect scaling.)

\paragraph{Selective reconstruction (per decode step).}
At decode step $t$, SR selects a head-specific index set $\mathcal{S}_{t,\ell,g}\subseteq[L]$ with
$M_{t,\ell,g}:=\lvert\mathcal{S}_{t,\ell,g}\rvert$ and reconstructs
$\widehat{\mathbf{X}}_{\ell,g}[\mathcal{S}_{t,\ell,g},:]
=
\mathbf{A}_k[\mathcal{S}_{t,\ell,g},:]\mathbf{B}_{\ell,g}$.
The per-layer SR reconstruction FLOPs are therefore
\begin{equation}
\label{eq:cost-sr}
\mathrm{FLOPs}_{\mathrm{SR}}
\;=\;
\sum_{g=1}^{H_{\mathrm{kv}}}
M_{t,\ell,g}\,d_h\,
\Bigl(
\underbrace{r_K}_{\text{keys}}
\;+\;
\underbrace{r_V}_{\text{values (if compressed)}}
\Bigr).
\end{equation}
When $M_{t,\ell,g}\ll L$, SR is a small fraction of the dense reconstruction cost.
The overhead for computing $\mathcal{S}_{t,\ell,g}$ (landmark scoring + Top-$k$) consists of lightweight
matrix--vector operations and is independent of the CLF factors.

\paragraph{Compressed-cache memory (CLF factors).}
Let $G := \lceil N/W\rceil$ be the number of windows.
CLF stores one shared basis per window and one reconstruction matrix per layer.
Thus, the total factor memory (counted as number of stored scalars) is
\begin{equation}
\label{eq:mem-factors}
M_{\mathrm{fact}}
\;=\;
\underbrace{\bigl(G\,L\,r_K + N\,r_K\,d\bigr)}_{\text{compressed keys}}
\;+\;
\underbrace{\bigl(G\,L\,r_V + N\,r_V\,d\bigr)}_{\text{compressed values (if compressed)}}.
\end{equation}
For reference, the uncompressed full KV-Cache stores
\begin{equation}
\label{eq:mem-full}
M_{\mathrm{full}} \;=\; 2N\,L\,d
\end{equation}
scalars (keys + values).

\paragraph{Landmark memory (optional, SR only).}
SR uses chunk landmarks built from keys with chunk size $c$ (Appendix~\ref{appendix: landmark detail}).
Let $n_c := \lceil L/c\rceil$ be the number of chunks.
Each layer stores $n_c\times d$ landmark scalars, so the total landmark memory is
\begin{equation}
\label{eq:mem-landmark}
M_{\mathrm{lm}}
\;=\;
N\,n_c\,d
\;\approx\;
\frac{N\,L\,d}{c}.
\end{equation}
Any static outlier indices require $O(NH_{\mathrm{kv}}o)$ integers and are negligible for long contexts.

\paragraph{How we compute KV-Cache compression ratios.}
We first define the \emph{factor-only} compression ratios for keys and values:
\begin{equation}
\label{eq:CK-CV}
C_K
\;:=\;
\frac{N\,L\,d}{G\,L\,r_K + N\,r_K\,d},
\qquad
C_V
\;:=\;
\frac{N\,L\,d}{G\,L\,r_V + N\,r_V\,d}.
\end{equation}
We also define the landmark fraction
\[
\alpha \;:=\; \frac{n_c}{L}\ \approx\ \frac{1}{c}
\qquad
(\text{e.g., }\alpha=\tfrac{1}{8}\text{ when }c=8).
\]
All ratios below are reported relative to the original combined KV size $2NLD$.

\emph{\xKSR{} (keys compressed, values offloaded).}
GPU memory usage consists of compressed keys plus landmarks, i.e.,
$NLD\cdot\bigl(\tfrac{1}{C_K}+\alpha\bigr)$, hence
\begin{equation}
\label{eq:ratio-xksr-gpu}
R_{\xKSR,\,\mathrm{GPU}}
\;=\;
\frac{2}{\tfrac{1}{C_K}+\alpha}.
\end{equation}
If counting \emph{total} memory with values stored at full size, we add $NLD$:
\begin{equation}
\label{eq:ratio-xksr-total}
R_{\xKSR,\,\mathrm{total}}
\;=\;
\frac{2}{\tfrac{1}{C_K}+\alpha+1}.
\end{equation}

\emph{\xKVSR{} (both keys and values compressed).}
GPU memory usage is $NLD\cdot\bigl(\tfrac{1}{C_K}+\tfrac{1}{C_V}+\alpha\bigr)$, hence
\begin{equation}
\label{eq:ratio-xkvsr}
R_{\xKVSR}
\;=\;
\frac{2}{\tfrac{1}{C_K}+\tfrac{1}{C_V}+\alpha}.
\end{equation}
If $C_K=C_V=C$ and $c=8$ (so $\alpha=1/8$), this reduces to
$R_{\xKSR,\,\mathrm{GPU}}=\frac{2}{\tfrac{1}{C}+\tfrac{1}{8}}$ and
$R_{\xKVSR}=\frac{2}{\tfrac{2}{C}+\tfrac{1}{8}}$.

\section{Extending \xKV{} on Multi-head Latent Attention (MLA)}
\label{appendix: MLA}
\begin{figure}[ht]
    \centering
    \includegraphics[width=0.9\linewidth]{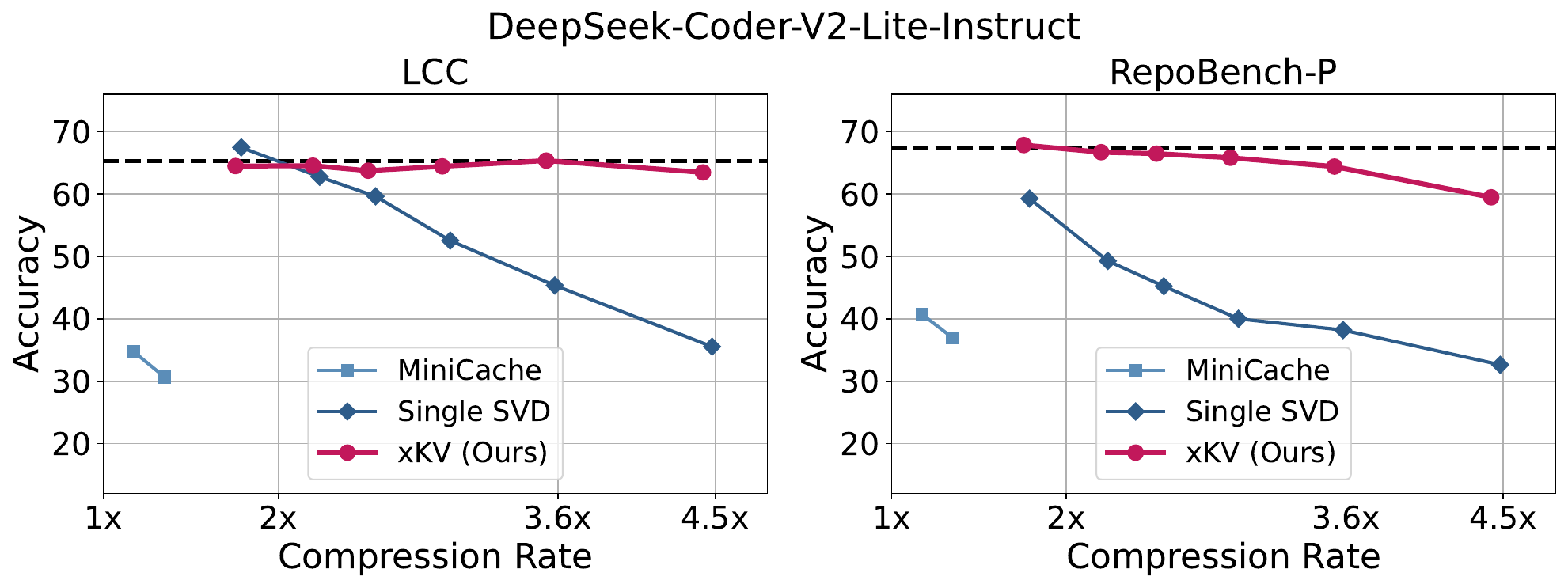}
    \label{fig:mla_lcc}
  \caption{Evaluation results of different \kv{} methods on DeepSeek-Coder-V2-Lite-Instruct model using RepoBench-P \citep{RepoBench} and LCC\citep{LCC}. The accuracy denotes the edit similarity \citep{svyatkovskiy2020intellicode}, and the dotted line represents the baseline score with uncompressed \kv{}.}
  \label{fig: mla}
\end{figure}
To demonstrate the effectiveness of \xKV{} on emerging attention variants, we evaluate xKV on DeepSeek-V2-Coder-Lite \citep{deepseekai2024deepseekv2}, which employs the efficient Multi-head Latent Attention (MLA) architecture~\citep{deepseekai2024deepseekv2}. MLA is proposed to reduce the \kv{} size per layer through low-rank projections. As shown in Figure~\ref{fig: mla}, we can further compress the compact latent cache by exploiting the cross-layer redundancy by using our \xKV{}. With a window size of 4, \xKV{} achieves a 3$\times$ compression rate on RepoBench \citep{RepoBench} and 3.5$\times$ on LCC \citep {LCC} without compromising accuracy. In contrast, other methods, such as MiniCache \citep{liu2024minicache} and Single SVD, fail to preserve accuracy on the MLA architecture even at substantially lower compression rates. 
These results underscore \xKV{}'s versatility and compatibility with emerging memory-efficient attention architectures \citep{deepseekai2024deepseekv2}.

\section{Broader CKA Analysis}
\label{appx: more cka}
We extend CKA analysis to a broader set of model architectures, including a small-scale dense model (Llama-3.2-1B) and a large-scale hybrid/MoE model (GPT-OSS 120B [1]), which features an interleaved 1:1 ratio of sliding-window and full attention layers.

Across these diverse settings, we observe that the characteristic singular-vector alignments are clearly preserved, strongly supporting the generality of our findings.

For the GPT-OSS model, we specifically noted that CKA similarity is highest between adjacent layers of the same attention type (e.g., Window→Window or Full→Full). This behavior suggests that xKV is naturally positioned for integration with future architectures that employ hybrid or interleaved attention designs (e.g., GPT-OSS \citep{openai2025gptoss120bgptoss20bmodel}, Kimi-Linear \citep{kimiteam2025kimilinearexpressiveefficient}).

\begin{figure}[ht]
    \centering

    \begin{subfigure}{0.5\linewidth}
        \centering
        \includegraphics[width=\linewidth]{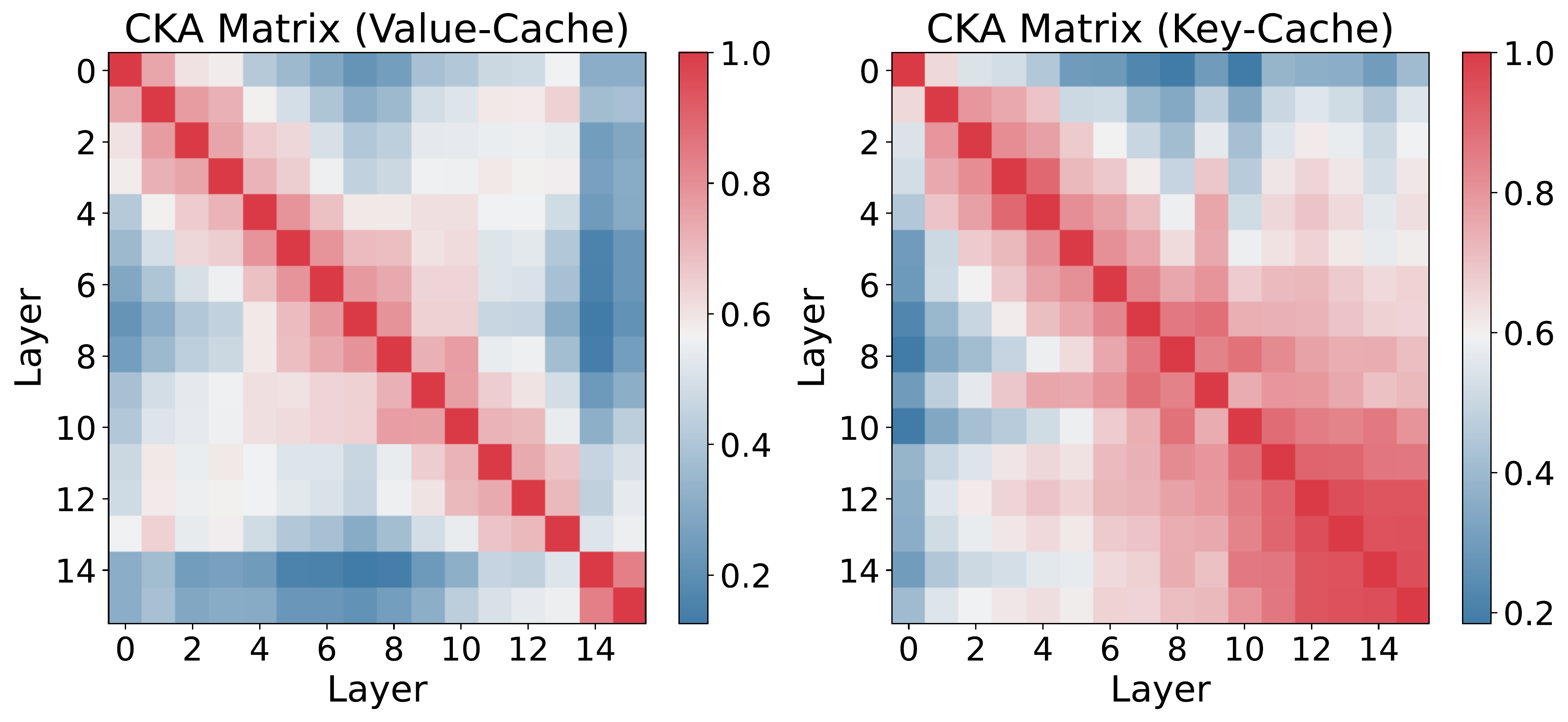}
        \caption{Llama3.2-1B}
        \label{fig:cka_llama}
    \end{subfigure}

    \begin{subfigure}{0.5\linewidth}
        \centering
        \includegraphics[width=\linewidth]{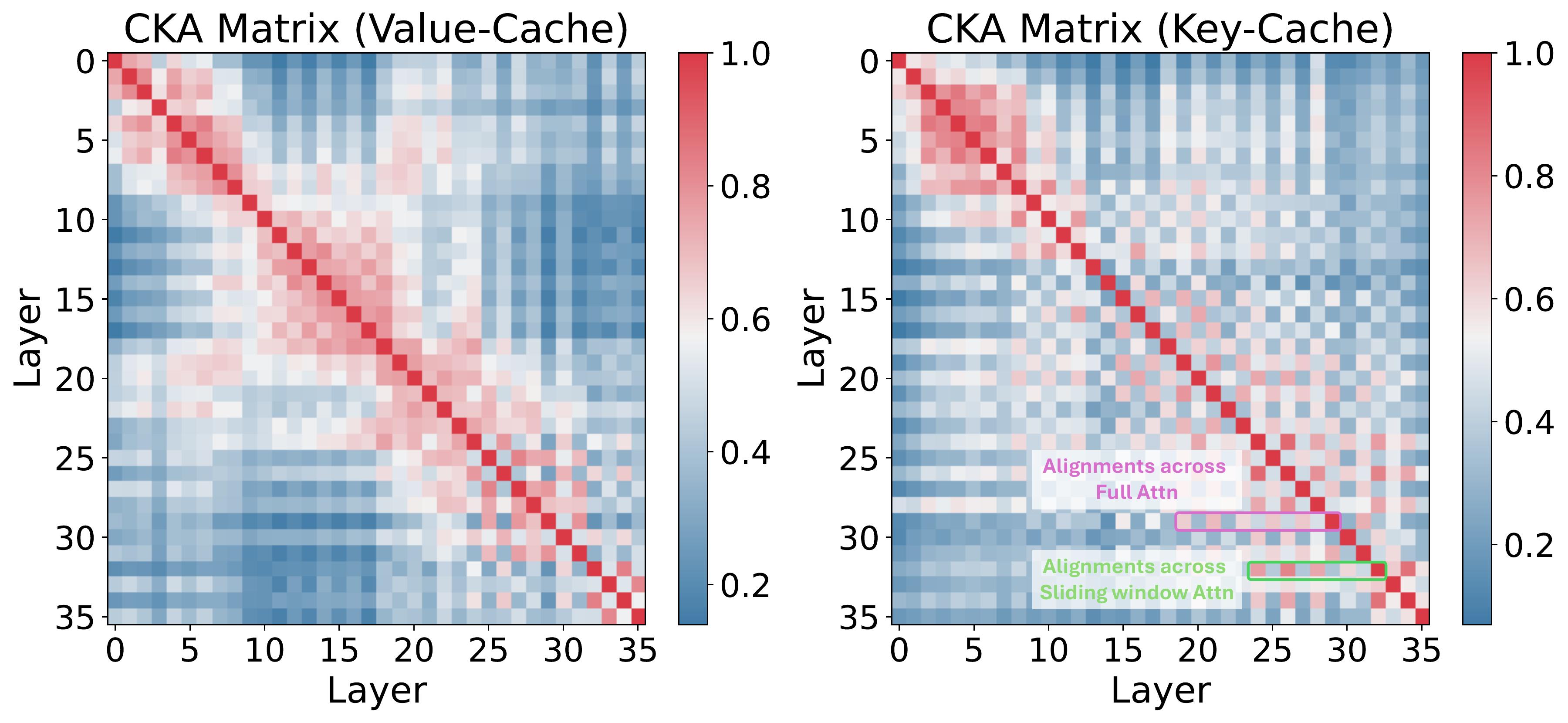}
        \caption{GPT-OSS 120B}
        \label{fig:cka_gptoss}
    \end{subfigure}

    \begin{subfigure}{0.5\linewidth}
        \centering
        \includegraphics[width=\linewidth]{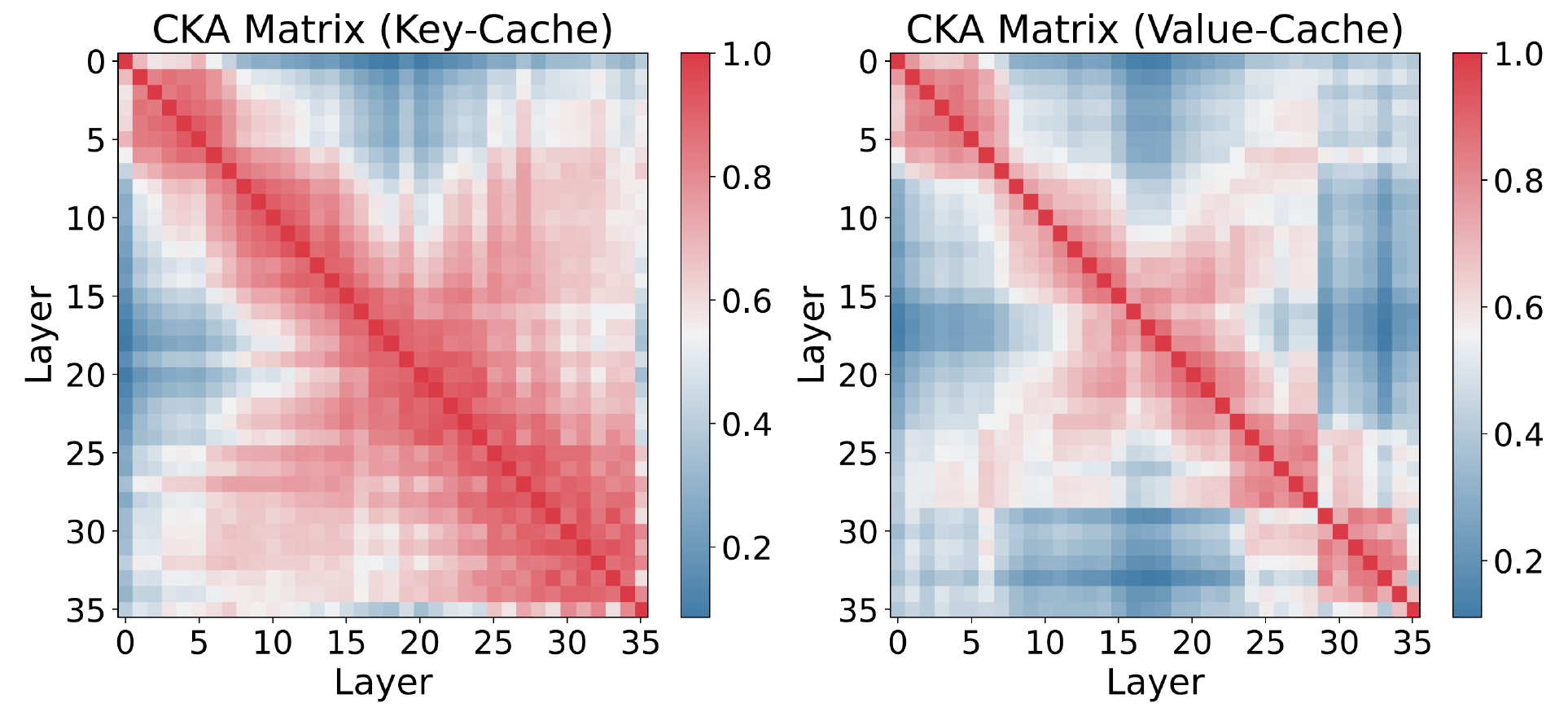}
        \caption{Qwen3-4B-Instruct}
        \label{fig:cka_new}
    \end{subfigure}

    \caption{Extended CKA analysis of three different models.}
    \label{fig:mla}
\end{figure}


\end{document}